%% file: Methods_and_algorithms.tex
\documentclass[10pt,twocolumn,twoside]{IEEEtran}
\usepackage{graphics, subfigure, theorem, times, amsfonts, amsmath, amssymb, cite, verbatim, math tools}
\usepackage{tikz}
\usepackage{framed}
\usetikzlibrary{shapes,arrows}
\input{figures/tikz_styles}
\usepackage{pgfplots}
\usepackage{color}
\input{my_sections.tex}

\usepackage{multirow}
\usepackage{rotating}
\usepackage{framed}
\input{mysymbol.sty}

\newtheorem{claim}{\hspace{0pt}\bf Claim}
\newtheorem{proposition}{\hspace{0pt}\bf Proposition}

\newtheorem{theorem}{\hspace{0pt}\bf Theorem}

\newtheorem{remark}{\hspace{0pt}\bf Remark}

\def\R {\text{\normalfont R}}

\def\NR {\text{\normalfont NR}}
\def\SR {\text{\normalfont SR}}
\def\SL {\text{\normalfont SL}}

\title{Admissible Hierarchical Clustering Methods and Algorithms for Asymmetric Networks}

\author{\IEEEauthorblockN{Gunnar Carlsson, Facundo M\'emoli, Alejandro Ribeiro, and Santiago Segarra}
\thanks{Authors are ordered alphabetically. Work in this paper is supported by NSF CCF-1217963, NSF CAREER CCF-0952867, NSF IIS-1422400, NSF CCF-1526513, AFOSR FA9550-09-0-1-0531, AFOSR FA9550-09-1-0643, NSF DMS-0905823, and NSF DMS-0406992. G. Carlsson is with the Dept. of Mathematics, Stanford University. F. M\'emoli is with the Dept. of Mathematics and the Dept. of Computer Science and Engineering, Ohio State University. A. Ribeiro and S. Segarra are with the Dept. of Electrical and Systems Engineering, University of Pennsylvania. Email: gunnar@math.stanford.edu, memoli@math.osu.edu, and \{aribeiro, ssegarra\}@seas.upenn.edu. Part of the results in this paper appeared in~\cite{Carlssonetal13_2}.}}

\begin{document}
\maketitle

\begin{abstract}%
This paper characterizes hierarchical clustering methods that abide by two previously introduced axioms -- thus, denominated admissible methods -- and proposes tractable algorithms for their implementation. We leverage the fact that, for asymmetric networks, every admissible method must be contained between reciprocal and nonreciprocal clustering, and describe three families of intermediate methods. Grafting methods exchange branches between dendrograms generated by different admissible methods. The convex combination family combines admissible methods through a convex operation in the space of dendrograms, and thirdly, the semi-reciprocal family clusters nodes that are related by strong cyclic influences in the network. Algorithms for the computation of hierarchical clusters generated by reciprocal and nonreciprocal clustering as well as the grafting, convex combination, and semi-reciprocal families are derived using matrix operations in a dioid algebra.
Finally, the introduced clustering methods and algorithms are exemplified through their application to a network describing the interrelation between sectors of the United States (U.S.) economy.
\end{abstract}


\section{Introduction} \label{sec_introduction}

The relevance of clustering in modern data analysis is indubitable given its usage in multiple fields of knowledge such as genetics \cite{Caronetal01}, computer vision \cite{Frigui99}, and sociology \cite{Handcock07}. There are literally hundreds of methods that can be applied to the determination of hierarchical \cite{lance67general, clusteringref} and non-hierarchical clusters in finite metric (thus symmetric) spaces -- see, e.g., \cite{RuiWunsch05}. Even in the case of asymmetric networks \cite{SaitoYadohisa04}, multiple methods have been developed to extend the notion of clustering into this less intuitive domain \cite{hubert-min,slater1976hierarchical,boyd-asymmetric,tarjan-improved,murtagh-multidimensional,PentneyMeila05}. Although not as developed as its practice \cite{science_art}, the theoretical framework for clustering has been developed over the last decade for non-hierarchical \cite{ben-david-reza, CarlssonMemoli10, kleinberg, VanLaarhoven14, Meila05, Meilaƒ2007873} and hierarchical clustering \cite{clust-um, Carlssonetal13, Carlssonetal13_3, Carlssonetal14}.
Of special interest to us is this last direction where it has been shown in \cite{clust-um} that single linkage \cite[Ch. 4]{clusteringref} is the unique hierarchical clustering method for finite metric spaces that satisfies three reasonable axiomatic statements.

Regarding hierarchical clustering of asymmetric networks, our work in \cite{Carlssonetal13} introduces the axioms of value -- in a network with two nodes, the nodes cluster together at resolutions at which both can influence each other -- and transformation -- reducing some pairwise dissimilarities and increasing none cannot increase the resolution at which clusters form -- as reasonable behaviors that we should expect to see in clustering methods. Although weak in appearance, these axioms lead to the stringent result that all methods that abide by them -- denominated \emph{admissible} methods -- must lie between two particular clustering methods in a well-defined sense. The first method, \emph{reciprocal clustering}, requires clusters to form through edges exhibiting low dissimilarity in both directions whereas the second method, \emph{nonreciprocal clustering}, allows clusters to form through cycles of small dissimilarity. When restricted to symmetric networks, reciprocal and nonreciprocal clustering yield equivalent outputs, which coincide with the output of single linkage.

The difference between reciprocal and nonreciprocal clustering for general asymmetric networks allows the existence of intermediate admissible methods. Hence, the contribution of this paper is twofold. First, we characterize intermediate clustering methods and study their properties. Second, we propose an algorithmic framework based on an alternative matrix dioid algebra to implement the intermediate methods introduced as well as reciprocal and nonreciprocal clustering.

In Section~\ref{sec_intermediate_clustering_methods} we unveil three families of intermediate clustering methods. The grafting methods consist of attaching the clustering output structures of the reciprocal and nonreciprocal methods in a way such that admissibility is guaranteed (Section \ref{sec_grafting}). We further present a construction that can be regarded as a convex combination in the space of clustering methods. This operation is shown to preserve admissibility therefore giving rise to a second family of admissible methods (Section \ref{sec_convex_comb}). A third family of admissible clustering methods is defined in the form of semi-reciprocal methods that allow the formation of cyclic influences in a more restrictive sense than nonreciprocal clustering but more permissive than reciprocal clustering (Section \ref{sec_inter_reciprocal}).

In Section~\ref{sec_algorithms}, we develop algorithms to compute the dendrograms associated with the methods introduced throughout the paper. The determination of algorithms for all of the methods introduced is given by the computation of matrix powers in a min-max dioid algebra \cite{GondranMinoux08}. In this algebra we operate in the field of positive reals and define the addition operation between two scalars to be their minimum and the product operation of two scalars to be their maximum. From this definition it follows that the $(i,j)$-th entry of the $l$-th dioid power of a matrix of network dissimilarities represents the minimax cost of a chain linking node $i$ to node $j$ with at most $l$ edges. Since reciprocal and nonreciprocal clustering require the determination of chains of minimax cost, their implementation can be framed in terms of dioid matrix powers. Similarly, other clustering methods introduced in this paper can be interpreted as minimax chain costs of a previously modified matrix of dissimilarities.

Clustering methods are exemplified through their application to a real-world network representing the interactions between economic sectors of the U.S. economy (Section \ref{sec_numerical_results}). The purpose of this application is to understand which information can be extracted by performing hierarchical clustering analyses based on the different methods proposed. While the bidirectional influence required for cluster formation in reciprocal clustering might be too restrictive, nonreciprocal clustering propagates influence through arbitrarily large cycles, a feature which might be undesirable in practice. An intermediate behavior can be obtained by utilizing semi-reciprocal clustering where the cyclic propagation of influence is closer to the real behavior of sectors within the economy and, thus, we obtain a more reasonable clustering output. Concluding remarks in Section~\ref{sec_conclusion} close the paper.

\section {Preliminaries}\label{sec_preliminaries}

We define a network $N=(X,A_X)$ as a set of $n$ points or nodes $X$ jointly specified with a real-valued dissimilarity function $A_X:X\times X \rightarrow \reals_+$. Dissimilarities $A_X(x,x')$ from $x$ to  $x'$ are non-negative, and null if and only if $x=x'$, but may not satisfy the triangle inequality and may be asymmetric, i.e. $A_X(x,x') \neq A_X(x',x)$ for some $x, x' \in X$. The values $A_X(x,x')$ can be grouped in a matrix which, as it does not lead to confusion, we also denote by $A_X\in\reals^{n\times n}$. A hierarchical clustering of the network $N=(X,A_X)$ is a dendrogram $D_X$ which by definition is a nested set of partitions $D_X(\delta)$ indexed by the resolution parameter $\delta\geq0$. Partitions in $D_X$ are such that for  $\delta=0$ each point $x$ is in a separate cluster, i.e., $D_X(0) = \big\{ \{x\}, \, x\in X\big\}$, and for some sufficiently coarse resolution $\delta_0$ all nodes are in the same cluster, i.e., $D_X(\delta_0) = \big\{ X \big\}$. The requirement of nested partitions means that if $x$ and $x'$ are in the same cluster at resolution $\delta$ they stay co-clustered for all larger resolutions $\delta' > \delta$. From these requirements it follows that dendrograms can be represented as trees \cite{clust-um}; see, e.g., Fig.~\ref{fig_reciprocal_example_io}-(a). When $x$ and $x'$ are co-clustered at resolution $\delta$ in $D_X$ we say that they are equivalent at that resolution and write $x\sim_{D_X(\delta)} x'$. 

An ultrametric $u_X:X\times X \rightarrow \reals_+$ on the set $X$ is a function that satisfies the symmetry $u_X(x, x')=u_X(x', x)$ and identity $u_X(x, x')=0 \iff x=x'$ properties as well as the strong triangle inequality 
\begin{equation}\label{eqn_strong_triangle_inequality}
    u_X(x,x') \leq \max \big(u_X(x,x''),  u_X(x'',x') \big),
\end{equation}
for all $x, x', x'' \in X$.
For a given dendrogram $D_X$ consider the minimum resolution $\delta$ at which $x$ and $x'$ are clustered together and define 
\begin{equation}\label{eqn_theo_dendrograms_as_ultrametrics_10}
   u_X(x,x') := \min \big\{ \delta\geq 0 \,|\, x\sim_{D_X(\delta)} x' \big\}.
\end{equation}
It can be shown that the function $u_X$ in \eqref{eqn_theo_dendrograms_as_ultrametrics_10} satisfies \eqref{eqn_strong_triangle_inequality} proving an equivalence between dendrograms and finite ultrametrics, \cite[Theorem 9]{clust-um}. While dendrograms are useful graphical representations, ultrametrics are more convenient to present the results contained in this paper.

In the description of hierarchical clustering methods the concepts of chain and chain cost are important. Given a network $(X, A_X)$ and $x, x' \in X$, a chain from $x$ to $x'$ is any \emph{ordered} sequence of nodes $[x=x_0, \ldots, x_{l-1}, x_{l}=x']$ starting at $x$ and finishing at $x'$. We use the notation $C(x, x')$ to denote one such chain. We define the cost of a chain as the maximum dissimilarity encountered when traversing its links in order. Thus, the directed minimum chain cost $\tdu^*_X(x, x')$ between $x$ and $x'$ is then defined as the minimum cost among all chains connecting $x$ to $x'$,
\begin{equation}\label{eqn_nonreciprocal_chains}
       \tdu^{*}_X(x, x') := \min_{C(x,x')} \,\,
                         \max_{i | x_i\in C(x,x')} A_X(x_i,x_{i+1}).
\end{equation} 

A hierarchical clustering method is a map $\ccalH:\ccalN \to \ccalD$ from the set of networks $\ccalN$ to the set of dendrograms $\ccalD$, or, equivalently, a map $\ccalH:\ccalN \to \ccalU$ mapping each network $N$ into the set $\ccalU$ of networks with ultrametrics as dissimilarity functions, i.e., $\ccalH(N)=(X, u_X)$. Our goal is to find methods $\ccalH$ that satisfy the following intuitive restrictions:

\myparagraph{(A1) Axiom of Value} Given a two-node network $N=(\{p,q\},A_{p,q})$ with $A_{p,q}(p,q)=\alpha$, and  $A_{p,q}(q,p)=\beta$, the ultrametric $(X,u_{p,q})=\ccalH(N)$ output by $\ccalH$ satisfies
\begin{equation}\label{eqn_two_node_network_ultrametric}
   u_{p,q}(p,q) = \max(\alpha,\beta).
\end{equation}
\myparagraph{(A2) Axiom of Transformation} Given networks $N_X=(X,A_X)$ and $N_Y=(Y,A_Y)$ and a dissimilarity reducing map $\phi:X\to Y$, i.e. a map $\phi$ such that for all $x,x' \in X$ it holds $A_X(x,x')\geq A_Y(\phi(x),\phi(x'))$, the outputs $(X,u_X)=\ccalH(N_X)$ and $(Y,u_Y)=\ccalH(N_Y)$ satisfy 
\begin{equation}\label{eqn_dissimilarity_reducing_ultrametric}
    u_X(x,x') \geq u_Y(\phi(x),\phi(x')).
\end{equation} 

\medskip\noindent We say that node $x$ is able to influence node $x'$ at resolution $\delta$ if the dissimilarity from $x$ to $x'$ is not greater than $\delta$. In two-node networks, our intuition dictates that a cluster is formed if nodes $p$ and $q$ are able to influence each other. Thus, axiom (A1) states that in a network with two nodes, the dendrogram $D_X$ has them merging at the maximum value of the two dissimilarities between them. 
Axiom (A2) captures the intuition that if a network is transformed such that some nodes become more similar but no pair of nodes increases its dissimilarity, then the transformed network should cluster at lower resolutions than the original one. Formally, (A2) states that a contraction of the dissimilarity function $A_X$ entails a contraction of the associated ultrametric $u_X$.

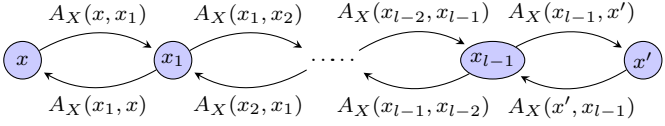
\begin{figure}
\centering
\centerline{\input{figures/reciprocal_path.tex} }
\vspace{-0.1in}
\caption{Reciprocal clustering. Nodes $x,x'$ cluster at resolution $\delta$ if they can be joined with a bidirectional chain of maximum dissimilarity $\delta$ [cf. \eqref{eqn_reciprocal_clustering}].}
\vspace{-0.05in}
\label{fig_reciprocal_path}
\end{figure}

A hierarchical clustering method $\ccalH$ is \emph{admissible} if it satisfies axioms (A1) and (A2). Two admissible methods of interest are reciprocal and nonreciprocal clustering. The \emph{reciprocal} clustering method $\ccalH^{\R}$ with output $(X,u^{\R}_X)=\ccalH^{\R}(X,A_X)$ is the one for which the ultrametric $u^{\R}_X(x,x')$ between points $x$ and $x'$ is given by
\begin{align}\label{eqn_reciprocal_clustering} 
    u^{\R}_X(x,x')
        &:= \min_{C(x,x')} \, \max_{i | x_i\in C(x,x')}
              \bbarA_X(x_i,x_{i+1}),
\end{align}
where $\bbarA_X(x,x'):=\max(A_X(x,x'), A_X(x',x))$. Definition \eqref{eqn_reciprocal_clustering} is illustrated in Fig. \ref{fig_reciprocal_path}. Intuitively, we search for chains $C(x, x')$ linking nodes $x$ and $x'$. For a given chain we walk from $x$ to $x'$ and for every link, connecting say $x_i$ with $x_{i+1}$, we determine the maximum dissimilarity in both directions, i.e. the value of $\bar{A}_X(x_i, x_{i+1})$. We then determine the maximum across all the links in the chain. The reciprocal ultrametric $u^{\R}_X(x, x')$ between $x$ and $x'$ is the minimum of this value across all possible chains.

Reciprocal clustering joins $x$ to $x'$ by going back and forth at maximum cost $\delta$ through the same chain. \emph{Nonreciprocal} clustering $\ccalH^{\NR}$ permits different chains and is defined as the maximum of the two minimum directed costs [cf.~\eqref{eqn_nonreciprocal_chains}] from $x$ to $x'$ and $x'$ to $x$
\begin{align}\label{eqn_nonreciprocal_clustering} 
    u^{\NR}_X(x,x') := \max \Big( \tdu^{*}_X(x,x'),\ \tdu^{*}_X(x',x )\Big).
\end{align}

%

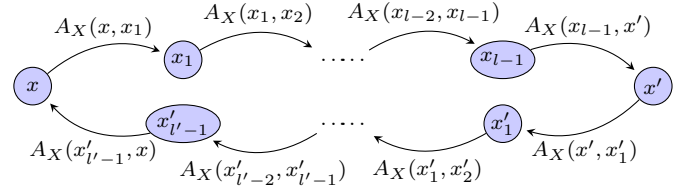
\begin{figure}
\centering
\input{figures/nonreciprocal_path.tex}
\vspace{-0.2in}
\caption{Nonreciprocal clustering. Nodes $x, x'$ cluster at resolution $\delta$ if they can be joined in both directions with possibly different chains of maximum dissimilarity $\delta$ [cf. \eqref{eqn_nonreciprocal_clustering}].}
\vspace{-0.05in}
\label{fig_nonreciprocal_path}
\end{figure}

\noindent Definition \eqref{eqn_nonreciprocal_clustering} is illustrated in Fig. \ref{fig_nonreciprocal_path}. We consider forward chains $C(x,x')$ going from $x$ to $x'$ and backward chains $C(x',x)$ going from $x'$ to $x$. We then determine the respective maximum dissimilarities and search independently for the best forward and backward chains that minimize these maximum dissimilarities. The nonreciprocal ultrametric $u^{\NR}_X(x,x')$ is the maximum of these two minimum values. Observe that since reciprocal chains are particular cases of nonreciprocal chains we must have $ u^{\NR}_X(x,x') \leq  u^{\R}_X(x,x')$ for all pairs of nodes $x,x' \in X$.

Reciprocal and nonreciprocal clustering are of importance because they bound the range of ultrametrics generated by any other admissible method $\ccalH$ in the sense stated next.

%
\begin{theorem}[\!\!\cite{Carlssonetal13}]\label{theo_extremal_ultrametrics} Consider an arbitrary network $N=(X,A_X)$ and let $u^{\R}_X$ and $u^{\NR}_X$ be the associated reciprocal and nonreciprocal ultrametrics as defined in \eqref{eqn_reciprocal_clustering} and \eqref{eqn_nonreciprocal_clustering}. Then, for any admissible method $\ccalH$ the output ultrametric $(X,u_X)=\ccalH(X,A_X)$ is such that for all pairs $x,x'$,
\begin{equation}\label{eqn_theo_extremal_ultrametrics} 
    u^{\NR}_X(x,x') \leq  u_X(x,x') \leq  u^{\R}_X(x,x').
\end{equation} 
In particular, $u^{\NR}_X=u^{\R}_X$ whenever $(X, A_X)$ is symmetric.
\end{theorem}

\smallskip\noindent According to Theorem 1, nonreciprocal clustering yields uniformly minimal ultrametrics while reciprocal clustering yields uniformly maximal ultrametrics among all methods satisfying (A1)-(A2). For symmetric networks, reciprocal and nonreciprocal clustering coincide, implying that there is a unique admissible method which is equivalent to the well-known single linkage hierarchical clustering method \cite[Ch. 4]{clusteringref}.

\section{Intermediate Clustering Methods}\label{sec_intermediate_clustering_methods}

Reciprocal and nonreciprocal clustering bound the range of methods satisfying axioms (A1)-(A2) in the sense specified by Theorem \ref{theo_extremal_ultrametrics}. Since $\ccalH^{\R}$ and $\ccalH^{\NR}$ are in general different, a question of interest is whether one can identify methods which are \emph{intermediate} to $\ccalH^{\R}$ and $\ccalH^{\NR}$.
We present three types of intermediate clustering methods: grafting, convex combinations, and semi-reciprocal clustering. The latter arises as a natural intermediate method in an algorithmic sense, as further discussed in Section~\ref{sec_algorithms}.

%

\subsection{Grafting}\label{sec_grafting}
\begin{figure}
\centering
\input{figures/reciprocal_nonreciprocal_dendrograms_2.tex}
\vspace{-0.1in}
\caption{Dendrogram grafting. Reciprocal ($\ccalH^{\R}$), nonreciprocal ($\ccalH^{\NR}$), and grafting ($\ccalH^{\R/\NR}(\beta=4)$) dendrograms for the given network are shown -- edges not drawn have dissimilarities greater than 5. To form the latter, branches of the reciprocal dendrogram are cut at resolution $\beta=4$ and replaced by the corresponding branches of the nonreciprocal dendrogram.}
\vspace{-0.1in}
\label{fig_beta_ultrametrics_2}
\end{figure}
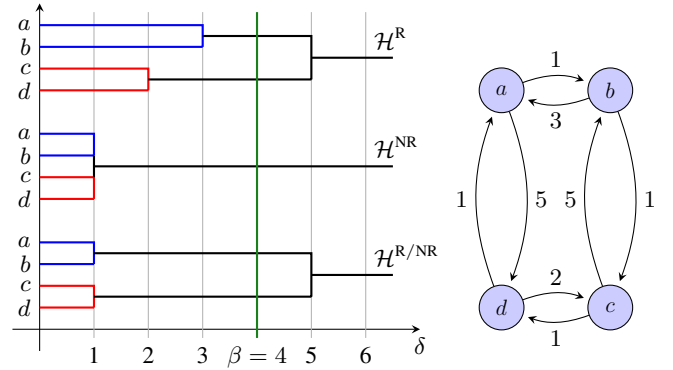

A family of admissible methods can be constructed by grafting branches of the nonreciprocal dendrogram into corresponding branches of the reciprocal dendrogram; see Fig. \ref{fig_beta_ultrametrics_2}. To be precise, consider a given positive constant $\beta>0$. For any given network $N=(X,A_X)$ compute the reciprocal and nonreciprocal dendrograms and cut all branches of the reciprocal dendrogram at resolution $\beta$. For each of these branches define the corresponding branch in the nonreciprocal tree as the one whose leaves are the same. Replacing the previously cut branches of the reciprocal tree by the corresponding branches of the nonreciprocal tree yields the $\ccalH^{\R/\NR}(\beta)$ method. Grafting is equivalent to providing the following piecewise definition of the output ultrametric
\begin{equation}\label{def_mu_beta_1}
   u^{\R/\NR}_X(x,x';\beta) :=
   \begin{cases}
      u^{\NR}_X(x,x'), & \text{if } u^{\R}_X(x,x') \leq \beta, \\
      u^{\R}_X(x,x'),  & \text{if } u^{\R}_X(x,x') >    \beta .
   \end{cases}
\end{equation}
For pairs $x,x' \in X$ having large reciprocal ultrametric value we keep this value, whereas for pairs with small reciprocal ultrametric value, we replace it by the nonreciprocal one.
 
To prove admissibility, we need to show that \eqref{def_mu_beta_1} defines an ultrametric and that the method $\ccalH^{\R/\NR}(\beta)$ satisfies axioms (A1) and (A2). This is asserted in the following proposition. 

\begin{proposition}\label{prop_beta_1}
The hierarchical clustering method $\ccalH^{\R/\NR}(\beta)$ is valid and admissible. I.e., $u_X^{\R/\NR}(\beta)$ defined in \eqref{def_mu_beta_1} is a valid ultrametric and $\ccalH^{\R/\NR}(\beta)$ satisfies axioms (A1)-(A2).
\end{proposition}
\begin{myproofnoname} 
The function $u^{\R/\NR}_X(\beta)$ fulfills the symmetry and identity properties of ultrametrics because $u^{\NR}_X$ and $u^{\R}_X$ fulfill them separately. Hence, to show that $u^{\R/\NR}_X(\beta)$ is a properly defined ultrametric, we need to show that it satisfies the strong triangle inequality (\ref{eqn_strong_triangle_inequality}).
%
 %
To show this, we split the proof into two cases: $u^{\R}_X(x,x') \leq \beta$ and $u^{\R}_X(x,x') > \beta$. Note that, by definition \eqref{def_mu_beta_1},
\begin{equation}\label{inter_beta_1}
u^{\NR}_X(x,x') \leq u^{\R/\NR}_X(x,x';\beta) \leq u^{\R}_X(x,x').
\end{equation}
Starting with the case where $u^{\R}_X(x,x') \leq \beta$, since $u^{\NR}_X$ satisfies (\ref{eqn_strong_triangle_inequality}) we can state that,
\begin{align}\label{beta_1_1}
u^{\R/\NR}_X(x,x';\beta)=& \, u^{\NR}_X(x,x') \nonumber\\
 \leq & \max \Big( u^{\NR}_X(x,x'') \, , \, u^{\NR}_X(x'',x')\Big).
\end{align}
Using the lower bound inequality in (\ref{inter_beta_1}) we can write 
\begin{align}\label{beta_1_2}
\max \Big( u^{\NR}_X&(x,x'') \, , \, u^{\NR}_X(x'',x')\Big) \nonumber \\
 \leq &\max \Big( u^{\R/\NR}_X(x,x'';\beta) \, , \, u^{\R/\NR}_X(x'',x';\beta)\Big).
\end{align}
Combining (\ref{beta_1_1}) and (\ref{beta_1_2}), we see that $u^{\R/\NR}_X(\beta)$ fulfills the strong triangle inequality in this case. As a second case, suppose that $u^{\R}_X(x,x') > \beta$, from the validity of the strong triangle inequality (\ref{eqn_strong_triangle_inequality}) for $u^{\R}_X$, we can write 
\begin{align}\label{beta_1_3}
\beta<u^{\R/\NR}_X(x,x';\beta)&=u^{\R}_X(x,x') \nonumber \\
&\leq \max \Big( u^{\R}_X(x,x'') \, , \, u^{\R}_X(x'',x')\Big).
\end{align}
This implies that at least one of $u^{\R}_X(x,x'')$ and $u^{\R}_X(x'',x')$ is greater than $\beta$. When this occurs, $u^{\R/\NR}_X(\beta)=u^{\R}_X$. Hence, 
\begin{align}\label{beta_1_4}
\max \Big( u^{\R}_X&(x,x'') \, , \, u^{\R}_X(x'',x')\Big) \nonumber \\
= \max &\Big( u^{\R/\NR}_X(x,x'';\beta) \, , \, u^{\R/\NR}_X(x'',x';\beta) \Big).
\end{align}
By substituting (\ref{beta_1_4}) into (\ref{beta_1_3}), we see that for this second case the strong triangle inequality is also satisfied.

To show that $\ccalH^{\R/\NR}(\beta)$ satisfies Axiom (A1) it suffices to see that in a two-node network $u^{\NR}_X$ and $u^{\R}_X$ coincide, meaning that we must have $u^{\R/\NR}_X(\beta) = u^{\NR}_X = u^{\R}_X$. Since $\ccalH^{\R}$ and $\ccalH^{\NR}$ fulfill (A1), the method $\ccalH^{\R/\NR}(\beta)$ must satisfy (A1) as well. 

To prove (A2) consider a dissimilarity reducing map $\phi:X\to Y$ and split consideration with regards to whether the reciprocal ultrametric is $u^{\R}_X(x,x') \leq \beta$ or $u^{\R}_X(x,x') > \beta$. When $u^{\R}_X(x,x') \leq \beta$ we must have $u^{\R}_Y(\phi(x),\phi(x')) \leq \beta$ because $\ccalH^{\R}$ satisfies (A2) and $\phi$ is a dissimilarity reducing map. Hence, according to the definition in (\ref{def_mu_beta_1}) we must have that both $u^{\R/\NR}_X(x,x';\beta)$ and $u^{\R/\NR}_Y(\phi(x),\phi(x');\beta)$ coincide with the nonreciprocal ultrametric and, since $\ccalH^{\NR}$ satisfies (A2), it immediately follows that $u^{\R/\NR}_X(x,x';\beta) \geq u^{\R/\NR}_Y(\phi(x),\phi(x');\beta)$, showing that $\ccalH^{\R/\NR}(\beta)$ satisfies (A2) when $u^{\R}_X(x,x') \leq \beta$. 

In the second case, when $u^{\R}_X(x,x') > \beta$, the validity of (A2) for the reciprocal ultrametric $u^{\R}_X$ allows us to write
\begin{equation}\label{beta_1_7}
u^{\R/\NR}_X(x,x';\beta)=u^{\R}_X(x,x') \geq u^{\R}_Y(\phi(x),\phi(x')).
\end{equation}
Combining this with the fact that $u^{\R}_Y$ is an upper bound on $u^{\R/\NR}_Y(\beta)$ [cf.~\eqref{inter_beta_1}], we see that $\ccalH^{\R/\NR}(\beta)$ satisfies (A2) also for this second case.
\end{myproofnoname}

Notice that, since $u^{\R/\NR}_X(x,x';\beta)$ coincides with either $u^{\NR}_X(x,x')$ or $u^{\R}_X(x,x')$ for all $x, x' \in X$, it satisfies Theorem \ref{theo_extremal_ultrametrics} as it should be the case for any admissible method.

An example implementation of $\ccalH^{\R/\NR}(\beta \! = \! 4)$ for a particular network is illustrated in Fig. \ref{fig_beta_ultrametrics_2}. The nonreciprocal ultrametric \eqref{eqn_nonreciprocal_clustering} is $u^{\NR}_X(x, x') = 1$ for all $x \neq x'$ due to the outmost clockwise loop visiting all nodes at cost 1. This is represented in the nonreciprocal $\ccalH^{\NR}$ dendrogram in Fig. \ref{fig_beta_ultrametrics_2}. For the reciprocal ultrametric \eqref{eqn_reciprocal_clustering} nodes $c$ and $d$ merge at resolution $u^\R_X(c, d)=2$, nodes $a$ and $b$ at resolution $u^\R_X(a,b)=3$, and they all join together at resolution $\delta=5$. This can be seen in the reciprocal $\ccalH^{\R}$ dendrogram. To determine $u^{\R/\NR}_X(x,x';4)$ use the piecewise definition in \eqref{def_mu_beta_1}. Since the reciprocal ultrametrics $u^\R_X(c, d)=2$ and $u^\R_X(a,b)=3$ are smaller than $\beta\!=\!4$ we set the grafted outcomes to the nonreciprocal ultrametrics to obtain $u^{\R/\NR}_X(c, d)=u^{\NR}_X(c, d)=1$ and $u^{\R/\NR}_X(a,b)=u^{\NR}_X(a,b)=1$. Since the remaining ultrametrics are $u^{\R}_X(x,x') = 5$ which exceed $\beta$ we set $u^{\R/\NR}_X(x,x';4) = u^{\R}_X(x,x') = 5$. This yields the $\ccalH^{\R/\NR}$ dendrogram in Fig. \ref{fig_beta_ultrametrics_2} which we interpret as cutting branches from $\ccalH^{\R}$ that we replace by the corresponding branches of~$\ccalH^{\NR}$.

In the method $\ccalH^{\R/\NR}(\beta)$ we use the reciprocal ultrametric as a decision variable in the piecewise definition \eqref{def_mu_beta_1} and use nonreciprocal ultrametrics for nodes having small reciprocal ultrametrics. There are three other possible grafting combinations $\ccalH^{\R/\R}(\beta)$, $\ccalH^{\NR/\R}(\beta)$ and $\ccalH^{\NR/\NR}(\beta)$ depending on which ultrametric is used as decision variable to swap branches and which of the two ultrametrics is used for nodes having small values of the decision ultrametric. E.g., in $\ccalH^{\R/\R}(\beta)$, we use reciprocal ultrametrics as decision variables and as the choice for small values of reciprocal ultrametrics,
\begin{equation}\label{def_mu_beta_4}
   u^{\R/\R}_X(x,x';\beta) :=
      \begin{cases}
         u^{\R}_X(x,x'), & \text{if } u^{\R}_X(x,x')\leq \beta, \\
         u^{\NR}_X(x,x'), & \text{if } u^{\R}_X(x,x')>\beta.
      \end{cases}
\end{equation}
However, the method $\ccalH^{\R/\R}(\beta)$ is \emph{not} valid because for some networks the function $u^{\R/\R}_X(\beta)$ is not an ultrametric as it violates the strong triangle inequality in \eqref{eqn_strong_triangle_inequality}. As a counterexample consider again the network in Fig. \ref{fig_beta_ultrametrics_2}. Applying the definition in \eqref{def_mu_beta_4} we obtain that $u^{\R/\R}_X(a,b;4) \!= \! u^{\R}_X(a,b) \! = \! 3$ while $u^{\R/\R}_X(a,c;4) \! = \! u^{\NR}_X(a,c) \! = \! 1$ and similarly $u^{\R/\R}_X(c,b;4) \! = \! 1$. In turn, this implies that $u^{\R/\R}_X(a,b;4) \! > \! \max(u^{\R/\R}_X(a,c;4), u^{\R/\R}_X(c,b;4))$ violating the strong triangle inequality. Analogously, $\ccalH^{\NR/\NR}(\beta)$ and $\ccalH^{\NR/\R}(\beta)$ can also be shown to be invalid clustering methods.

A second valid grafting alternative can be obtained as a modification of $\ccalH^{\R/\R}(\beta)$ in which reciprocal ultrametrics are kept for pairs having small reciprocal ultrametrics, nonreciprocal ultrametrics are used for pairs having large reciprocal ultrametrics, but all nonreciprocal ultrametrics smaller than $\beta$ are saturated to this value. Denoting the method by $\ccalH^{\R/\R_{\max}}(\beta)$ the output ultrametrics are thereby given as
\begin{equation}\label{def_mu_beta_5}
   u^{\R/\R_{\max}}_X\!(x,x';\beta) :=
   \begin{cases}
      u^{\R}_X(x,x'),                        &\!\!\! \text{if } u^{\R}_X(x,x')\leq \beta, \\
      \max \big(\beta, u^{\NR}_X(x,x')\big), &\!\!\! \text{if } u^{\R}_X(x,x')>\beta.
   \end{cases}
\end{equation}
This alternative definition outputs a valid ultrametric and $\ccalH^{\R/\R_{\max}}(\beta)$ satisfies axioms (A1)-(A2) as claimed next.

\begin{proposition}\label{prop_beta_5}
The method $\ccalH^{\R/\R_{\max}} (\beta)$ is valid and admissible. I.e., $u_X^{\R/\R_{\max}}\!(\beta)$ defined in \eqref{def_mu_beta_5} is a valid ultrametric and $\ccalH^{\R/\R_{\max}}(\beta)$ satisfies axioms (A1)-(A2).
\end{proposition}
\begin{myproofnoname} 
This proof follows from a reasoning analogous to that in the proof of Proposition~\ref{prop_beta_1}. In particular, by definition we have that [cf.~\eqref{inter_beta_1}]
\begin{equation}\label{inter_beta_5}
u^{\NR}_X(x,x') \leq u^{\R/\R_{\max}}_X(x,x';\beta) \leq u^{\R}_X(x,x'),
\end{equation}
which immediately implies fulfillment of (A1). Also, as done for Proposition~\ref{prop_beta_1}, the strong triangle inequality and the fulfillment of (A2) can be shown by dividing the proofs into the two cases $u^{\R}_X(x,x') \leq \beta$ and $u^{\R}_X(x,x') > \beta$.
\end{myproofnoname}

\begin{remark}\normalfont
Intuitively, the grafting combination $\ccalH^{\R/\NR}(\beta)$ allows nonreciprocal propagation of influence for resolutions smaller than $\beta$ while requiring reciprocal propagation for higher resolutions. This is of interest if we want tight clusters of small dissimilarity to be formed through loops of influence while looser clusters of higher dissimilarity are required to form through links of bidirectional influence. Conversely, the clustering method $\ccalH^{\R/\R_{\max}}(\beta)$ requires reciprocal influence within tight clusters of resolution smaller than $\beta$ but allows nonreciprocal influence in clusters of higher resolutions. This latter behavior is desirable in, e.g., trust propagation in social interactions, where we want tight clusters to be formed through links of mutual trust but allow looser clusters to be formed through unidirectional trust loops.
\end{remark}

\subsection{Convex combinations}\label{sec_convex_comb}

A different family of intermediate admissible methods can be constructed by performing a convex combination of methods known to satisfy axioms (A1) and (A2). Indeed, consider two admissible clustering methods $\ccalH^1$ and $\ccalH^2$ and a given parameter $0 \leq \theta \leq 1$. For an arbitrary network $N=(X, A_X)$ denote by $(X,u^1_X) = \ccalH^1(N)$ and $(X,u^2_X)=\ccalH^2(N)$ the respective outcomes of methods $\ccalH^1$ and $\ccalH^2$. Construct then the dissimilarity function $A_X^{12}(\theta)$ as the convex combination of $u^1_X$ and $u^2_X$, for all $x,x'\in X$
\begin{equation}\label{eqn_def_sym_net_conv_comb}
   A^{12}_X(x,x';\theta) := \theta\, u^1_X(x, x') + (1-\theta) \, u^2_X(x, x').
\end{equation}
Although $A_X^{12}(\theta)$ is a well-defined dissimilarity function, it is not an ultrametric in general because it may violate the strong triangle inequality. Nevertheless, we can recover the ultrametric structure by applying any admissible clustering method $\ccalH$ to the symmetric network $N^{12}_\theta=(X,A^{12}_X(\theta))$. Moreover, as explained after Theorem~\ref{theo_extremal_ultrametrics}, single linkage is the unique admissible clustering method for symmetric networks. Thus, we define the convex combination method $\ccalH^{12}_\theta$ as the application of single linkage on $N^{12}_\theta$. Formally, we define $\ccalH^{12}_\theta$ as a method whose output $(X,u^{12}_{X}(\theta))=\ccalH^{12}_\theta(N)$ corresponding to network $N=(X,A_X)$ is given by
\begin{align}\label{eqn_def_u_1_2_bar}
   u^{12}_{X}(x, x';\theta) 
      := \min_{C(x,x')} \, \max_{i | x_i\in C(x,x')} A^{12}_X(x_i,x_{i+1};\theta),
\end{align}
for all $x, x' \in X$ and $A^{12}_X(\theta)$ as given in \eqref{eqn_def_sym_net_conv_comb}. We show that \eqref{eqn_def_u_1_2_bar} defines a valid ultrametric and that $\ccalH^{12}_\theta$ fulfills axioms (A1) and (A2) in the following proposition.

\begin{proposition}\label{prop_convex_combination}
Given two admissible hierarchical clustering methods $\ccalH^1$ and $\ccalH^2$, the convex combination method $\ccalH^{12}_\theta$ is valid and admissible. I.e., $u_X^{12}(\theta)$ defined in \eqref{eqn_def_u_1_2_bar} is a valid ultrametric and $\ccalH^{12}_\theta$ satisfies axioms (A1)-(A2).
\end{proposition}
\begin{myproofnoname}
As discussed in the paragraph preceding the statement of this proposition, $u^{12}_X(\theta)$ is the output of applying single linkage to the symmetric network $N^{12}_\theta$, immediately implying that $u^{12}_X(\theta)$ is a well-defined ultrametric.

To see that axiom (A1) is fulfilled, pick an arbitrary two-node network $(\{p,q\}, A_{p, q})$ with $A_{p, q}(p, q)=\alpha$ and $A_{p, q}(q, p)=\beta$. Since methods $\ccalH^1$ and $\ccalH^2$ are admissible, in particular they satisfy (A1), hence $u^1_{p, q}(p,q)=u^2_{p, q}(p,q)=\max(\alpha, \beta)$. It then follows from \eqref{eqn_def_sym_net_conv_comb} that $A^{12}_{p, q}(p,q;\theta)=\max(\alpha, \beta)$ for all possible values of $\theta$. Moreover, since in \eqref{eqn_def_u_1_2_bar} all possible chains joining $p$ and $q$ must contain these two nodes as consecutive elements, we have that
\begin{equation}
u^{12}_{p, q}(p,q;\theta)=A^{12}_{p, q}(p,q;\theta)=\max(\alpha, \beta),
\end{equation}
for all $\theta$, satisfying axiom (A1).

Fulfillment of axiom (A2) also follows from admissibility of $\ccalH^1$ and $\ccalH^2$. Suppose there are two networks $N_X=(X,A_X)$ and $N_Y=(Y, A_Y)$ and a dissimilarity reducing map $\phi: X \to Y$. From the facts that $\ccalH^1$ and $\ccalH^2$ satisfy (A2) we have
\begin{align}
u^1_X(x, x') \geq u^1_Y(\phi(x), \phi(x')), \,\,\, u^2_X(x, x') \geq u^2_Y(\phi(x), \phi(x')) \label{eqn_ultram_2_convex_comb}.
\end{align}
By multiplying the left inequality by $\theta$ and the right one by $(1-\theta)$, and adding both inequalities we obtain [cf. \eqref{eqn_def_sym_net_conv_comb}]
\begin{align}
A^{12}_X(x, x';\theta) \geq A^{12}_Y(\phi(x), \phi(x');\theta),
\end{align}
for all $0 \leq \theta \leq 1$. This implies that the map $\phi$ is also dissimilarity reducing between the networks $(X, A^{12}_X(\theta))$ and $(Y, A^{12}_Y(\theta))$. Combining this with the fact that we apply an admissible method (single linkage) to the previous networks to obtain the ultrametric outputs, it follows that
\begin{align}
u^{12}_X(x, x';\theta) \geq u^{12}_Y(\phi(x), \phi(x');\theta),
\end{align}
for all $\theta$, showing that axiom (A2) is satisfied by the convex combination method.
\end{myproofnoname}

The construction in \eqref{eqn_def_u_1_2_bar} can be generalized to produce intermediate clustering methods generated by convex combinations of any number (i.e. not necessarily two) of admissible methods. These convex combinations can be seen to satisfy axioms (A1) and (A2) through recursive applications of Proposition~\ref{prop_convex_combination}.

\vspace{-0.1in}
\begin{remark}\normalfont
Since \eqref{eqn_def_u_1_2_bar} is equivalent to single linkage applied to the symmetric network $N^{12}_\theta$, it follows \cite{clust-um,CarlssonMemoli10} that $u_X^{12}(\theta)$ is the largest ultrametric bounded above by $A^{12}_X(\theta)$, i.e., the largest ultrametric for which $u^{12}_X(x,x';\theta)\leq A^{12}_X(x,x';\theta)$ for all $x,x'$. We can then think of \eqref{eqn_def_u_1_2_bar} as an operation ensuring a valid ultrametric definition while deviating as little as possible from $A^{12}_X(\theta)$, thus, retaining as much information as possible in the convex combination of $u^1_X$ and $u^2_X$.
\end{remark}

%
\begin{figure}
\centering
\centerline{\input{figures/inter_reciprocal_example.tex}}
\vspace{-0.2in}
\caption{Semi-reciprocal chains. The main chain joining $x$ and $x'$ is formed by $[x, x_1, ... , x_{r}, x']$. Between two consecutive nodes of the main chain $x_i$ and $x_{i+1}$, we have a secondary chain in each direction. For $u^{\SR(t)}_X$, the maximum allowed node-length of secondary chains is $t$.}
\vspace{-0.1in}
\label{fig_inter_reciprocal_example}
\end{figure}
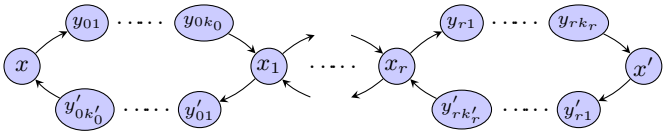

\subsection{Semi-reciprocal}\label{sec_inter_reciprocal}

In reciprocal clustering we require influence to propagate through bidirectional chains; see Fig. \ref{fig_reciprocal_path}. We could reinterpret bidirectional propagation as allowing loops of node-length two in both directions. E.g., the bidirectional chain between $x$ and $x_1$ in Fig. \ref{fig_reciprocal_path} can be interpreted as a loop between $x$ and $x_1$ composed by two chains $[x, x_1]$ and $[x_1, x]$ of node-length two. \emph{Semi-reciprocal} clustering is a generalization of this concept where loops consisting of at most $t$ nodes in each direction are allowed. Given $t \in \naturals$ such that $t \geq 2$, we use the notation $C_t(x,x')$ to denote any chain $[x=x_0,x_1,\ldots,x_l=x']$ joining $x$ to $x'$ where $l \leq t-1$. That is, $C_t(x,x')$ is a chain starting at $x$ and finishing at $x'$ with at most $t$ nodes. We reserve the notation $C(x,x')$ to represent a chain from $x$ to $x'$ where no maximum is imposed on the number of nodes. Given an arbitrary network $N=(X, A_X)$, define as $A^{\SR(t)}_X(x, x')$ the minimum cost incurred when traveling from node $x$ to node $x'$ using a chain of at most $t$ nodes. I.e.,
\begin{equation}\label{eqn_inter_cost}
A^{\SR(t)}_X(x, x'):=\min_{C_t(x, x')} \,\,\,  \max_{i | x_i\in C_t(x, x')} A_X(x_i, x_{i+1}).
\end{equation}
We define the family of semi-reciprocal clustering methods $\ccalH^{\SR(t)}$ with output $(X,u^{\SR(t)}_X)=\ccalH^{\SR(t)}(X,A_X)$ as the one for which the ultrametric $u^{\SR(t)}_X(x,x')$ between $x$ and $x'$ is 
\begin{align}\label{eqn_inter_reciprocal_clustering} 
    u^{\SR(t)}_X(x,x') := \min_{C(x,x')} \,\,\, \max_{i | x_i\in C(x,x')} \bar{A}^{\SR(t)}_X(x_i, x_{i+1}),
\end{align}     
where the function $\bar{A}^{\SR(t)}_X$ is defined as
\begin{align}\label{eqn_inter_reciprocal_clustering_auxiliary}     
    \bar{A}^{\SR(t)}_X(x_i, x_{i+1}) := \max \big(A^{\SR(t)}_X(x_i, x_{i+1}), A^{\SR(t)}_X(x_{i+1}, x_i)\big).
\end{align} 

The chain $C(x, x')$ of unconstrained length in \eqref{eqn_inter_reciprocal_clustering} is called the \emph{main chain}, represented by $[x=x_0, x_1, ... , x_{r}, x']$ in Fig. \ref{fig_inter_reciprocal_example}. Between consecutive nodes $x_i$ and $x_{i+1}$ of the main chain, we build loops consisting of secondary chains in each direction, represented in Fig. \ref{fig_inter_reciprocal_example} by $[x_i, y_{i1}, ... , y_{ik_i}, x_{i+1}]$ and $[x_{i+1}, y'_{i1}, ... , y'_{ik'_i}, x_{i}]$ for all $i$. 
For the computation of $u^{\SR(t)}_X(x,x')$, the maximum allowed length of secondary chains is equal to $t$ nodes, i.e., $k_i, k'_i \leq t-2$ for all $i$. In particular, for $t=2$ we recover the reciprocal chain; see Fig.~\ref{fig_reciprocal_path}.

We can reinterpret \eqref{eqn_inter_reciprocal_clustering} as the application of reciprocal clustering [cf. \eqref{eqn_reciprocal_clustering}] to a network with dissimilarities $A^{\SR(t)}_X$ as in \eqref{eqn_inter_cost}, i.e., a network with dissimilarities given by the optimal choice of secondary chains. Semi-reciprocal clustering methods are valid and satisfy axioms (A1)-(A2) as shown in the following proposition.
 
%
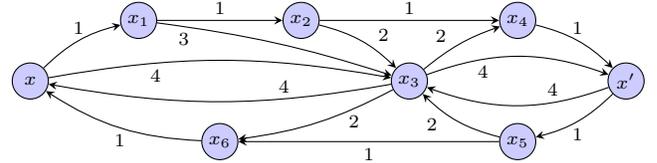
\begin{figure}
\centering
\centerline{\input{figures/inter_reciprocal_example_2.tex}}
\vspace{-0.1in}
\caption{Semi-reciprocal example. Computation of semi-reciprocal ultrametrics between nodes $x$ and $x'$ for different values of parameter $t$; see text for details.}
\vspace{-0.1in}
\label{fig_inter_reciprocal_example_2}
\end{figure}

\begin{proposition}\label{inter_reciprocal_axioms}
The semi-reciprocal clustering method $\ccalH^{\SR(t)}$ is valid and admissible for all integers $t \geq 2$. I.e., $u_X^{\SR(t)}$ is a valid ultrametric and $\ccalH^{\SR(t)}$ satisfies axioms (A1)-(A2).
\end{proposition}
\begin{myproofnoname}
We begin the proof by showing that \eqref{eqn_inter_reciprocal_clustering} outputs a valid ultrametric where the only non-trivial property to be shown is the strong triangle inequality \eqref{eqn_strong_triangle_inequality}. For a fixed $t$, pick an arbitrary pair of nodes $x$ and $x'$ and an arbitrary intermediate node $x''$. Let us denote by $C^*(x,x'')$ and $C^*(x'',x')$ a pair of main chains that satisfy definition \eqref{eqn_inter_reciprocal_clustering} for $u^{\SR(t)}_X(x,x'')$ and $u^{\SR(t)}_X(x'',x')$ respectively. Construct $C(x,x')$ by concatenating the aforementioned minimizing chains $C^*(x,x'')$ and $C^*(x'',x')$. However, $C(x,x')$ is a particular chain for computing $u^{\SR(t)}_X(x,x')$ and need not be the minimizing one. This implies that
\begin{equation}\label{eqn_triangle_inter_reciprocal}
u^{\SR(t)}_X(x,x') \leq \max \Big(u^{\SR(t)}_X(x,x''), u^{\SR(t)}_X(x'',x')\Big),
\end{equation}
proving the strong triangle inequality.

To show fulfillment of (A1), consider the network $(\{p,q\}, A_{p, q})$ with $A_{p, q}(p,q)=\alpha$ and $A_{p, q}(q,p)=\beta$. Note that in this situation, $A_{p, q}^{\SR(t)}(p, q)=\alpha$ and $A_{p, q}^{\SR(t)}(q, p)=\beta$ for all $t \geq 2$ [cf. \eqref{eqn_inter_cost}], since there is only one possible chain between them and contains only two nodes. Hence, from \eqref{eqn_inter_reciprocal_clustering},
\begin{equation}
u^{\SR(t)}_{p, q}(p,q) = \max (\alpha, \beta),
\end{equation}
for all $t$. Consequently, axiom (A1) is satisfied.

To show fulfillment of (A2), consider two arbitrary networks $(X, A_X)$ and $(Y, A_Y)$ and a dissimilarity reducing map $\phi: X \to Y$ between them. Further, denote by $C^*_X(x,x')=[x=x_0,\ldots, x_l =x']$ a main chain that achieves the minimum semi-reciprocal cost in \eqref{eqn_inter_reciprocal_clustering}. Then, for a fixed $t$, we can write
\begin{equation}\label{eqn_inter_reciprocal_axiom_3}
    u^{\SR(t)}_X(x,x') = \max_{i | x_i\in C^*_X(x,x')} \bar{A}^{\SR(t)}_X(x_i,x_{i+1}).
\end{equation}
Consider now a secondary chain $C^X_t(x_i, x_{i+1})=[x_i=x^{(0)},\ldots, x^{(l')}=x_{i+1}]$ between two consecutive nodes $x_i$ and $x_{i+1}$ of the minimizing chain $C^*_X(x,x')$. Further, focus on the image of this secondary chain under the map $\phi$, that is $C^Y_t(\phi(x_i),\phi(x_{i+1})):=\phi\big(C^X_t(x_i, x_{i+1})\big) = [\phi(x_i)=\phi(x^{(0)}),\ldots, \phi(x^{(l')})=\phi(x_{i+1})]$ in the set $Y$. 

Since the map $\phi:X\to Y$ is dissimilarity reducing, $A_Y(\phi(x^{(i)}),\phi(x^{(i+1)}))\leq A_X(x^{(i)},x^{(i+1)})$ for all links in this chain. Analogously, we can bound the dissimilarities in secondary chains $C^X_t(x_{i+1}, x_{i})$ from $x_{i+1}$ back to $x_i$. Thus, from \eqref{eqn_inter_cost} we can state that,
\begin{align}\label{eqn_inter_reciprocal_axiom_3_2}
    \bar{A}^{\SR(t)}_X(x_i,x_{i+1}) \geq \bar{A}^{\SR(t)}_Y(\phi(x_i),\phi(x_{i+1})).
\end{align}
Denote by $C_Y(\phi(x),\phi(x'))$ the image of the main chain $C^*_X(x,x')$ under the map $\phi$. Notice that $C_Y(\phi(x),\phi(x'))$ is a particular chain joining $\phi(x)$ and $\phi(x')$, whereas the semi-reciprocal ultrametric computes the minimum across all main chains. Therefore,
\begin{equation}\label{eqn_inter_reciprocal_axiom_3_3}
u^{\SR(t)}_Y\!(\phi(x),\phi(x'))  \leq  \!\!\! \max_{i | \phi(x_i) \in C_Y(\phi(x),\phi(x'))}  \!\!\!\!\!\!\! \bar{A}^{\SR(t)}_Y (\phi(x_i),\phi(x_{i+1}) \!).
\end{equation}
By bounding the right-hand side of \eqref{eqn_inter_reciprocal_axiom_3_3} using \eqref{eqn_inter_reciprocal_axiom_3_2} and recalling \eqref{eqn_inter_reciprocal_axiom_3}, it follows that $u^{\SR(t)}_Y(\phi(x),\phi(x'))\leq u^{\SR(t)}_X(x,x')$. This proves that (A2) is satisfied. 
\end{myproofnoname}

The semi-reciprocal family is a countable family of clustering methods parameterized by integer $t \geq 2$ representing the allowed maximum node-length of secondary chains. Reciprocal and nonreciprocal ultrametrics are equivalent to semi-reciprocal ultrametrics for specific values of $t$. For $t=2$ we have $u^{\SR(2)}_X = u^{\R}_X$ meaning that we recover reciprocal clustering. To see this formally, note that $A^{\SR(2)}_X(x, x')=A_X(x,x')$ [cf. \eqref{eqn_inter_cost}] since the only chain of length two joining $x$ and $x'$ is $[x, x']$. Hence, for $t=2$,  \eqref{eqn_inter_reciprocal_clustering} reduces to
\begin{equation}\label{eqn_inter_reciprocal_clustering_1} 
    u^{\SR(2)}_X(x,x')= \min_{C(x,x')} \,\,\, \max_{i | x_i\in C(x,x')} \bbarA_X(x_i,x_{i+1}),
\end{equation} 
which is the definition of the reciprocal ultrametric [cf. \eqref{eqn_reciprocal_clustering}]. Nonreciprocal ultrametrics can be obtained as $u^{\SR(t)}_X = u^{\NR}_X$ for any parameter $t$ exceeding the number of nodes in the network analyzed. To see this, notice that minimizing over $C(x, x')$ is equivalent to minimizing over $C_t(x, x')$ for all $t \geq n$, since we are looking for minimizing chains in a network with non-negative dissimilarities. Therefore, visiting the same node twice is not an optimal choice. This implies that $C_n(x,x')$ contains all possible minimizing chains between $x$ and $x'$. I.e., all chains of interest have at most $n$ nodes. Hence, by inspecting \eqref{eqn_inter_cost}, $A^{\SR(t)}_X(x, x')= \tdu^*_X(x, x')$ [cf. \eqref{eqn_nonreciprocal_chains}] for all $t \geq n$. Furthermore, when $t \geq n$, the best main chain that can be picked is formed only by nodes $x$ and $x'$ because, in this way, no additional meeting point is enforced between the chains going from $x$ to $x'$ and vice versa. As a consequence, definition \eqref{eqn_inter_reciprocal_clustering} reduces to
\begin{equation}\label{eqn_inter_reciprocal_clustering_2} 
    u^{\SR(t)}_X(x,x')= \max \Big(\tdu^*_X(x, x'), \tdu^*_X(x', x) \Big),
\end{equation} 
for all $x, x' \in X$ and for all $t \geq n$. The right hand side of \eqref{eqn_inter_reciprocal_clustering_2} is the definition of the nonreciprocal ultrametric [cf. \eqref{eqn_nonreciprocal_clustering}].

For the network in Fig. \ref{fig_inter_reciprocal_example_2}, we compute the semi-reciprocal ultrametrics between $x$ and $x'$ for different values of $t$. The edges which are not delineated are assigned dissimilarity values greater than $4$. Since the only bidirectional chain between $x$ and $x'$ uses $x_3$ as the intermediate node, we conclude that $u_X^{\R}(x,x')=u_X^{\SR(2)}(x,x')=4$. Furthermore, by constructing a path through the outermost clockwise cycle in the network, we conclude that $u_X^{\NR}(x, x')=1$. Since the longest secondary chain in the minimizing chain for the nonreciprocal case, $[x, x_1, x_2, x_4, x']$, has node-length 5, we may conclude that $u_X^{\SR(t)}(x, x')=1$ for all $t \geq 5$. For intermediate values of $t$, if e.g., we fix $t=3$, the minimizing chain is given by the main chain $[x, x_3, x']$ and the secondary chains $[x, x_1, x_3]$, $[x_3, x_4, x']$, $[x', x_5, x_3]$ and $[x_3, x_6, x]$ joining consecutive nodes in the main chain in both directions. The maximum cost among all dissimilarities in this path is $A_X(x_1, x_3)=3$. Hence, $u^{\SR(3)}_X(x, x')=3$. The minimizing chain for $t=4$ is similar to the minimizing one for $t=3$ but replacing the secondary chain $[x, x_1, x_3]$ by $[x, x_1, x_2, x_3]$. In this way, we obtain $u^{\SR(4)}_X(x, x')=2$.

\begin{remark}\normalfont 
Intuitively, when propagating influence through a network, reciprocal clustering requires bidirectional influence whereas nonreciprocal clustering allows arbitrarily large unidirectional cycles. In many applications, such as trust propagation in social networks, it is reasonable to look for an intermediate situation where influence can propagate through cycles but of limited length. Semi-reciprocal ultrametrics represent this intermediate situation where the parameter $t$ represents the maximum length of chains through which influence can propagate in a nonreciprocal manner.
\end{remark}

\section{Algorithms}\label{sec_algorithms}

Recall that, for convenience, we can interpret the dissimilarity function $A_X$ as an $n\times n$ matrix and, similarly, $u_X$ can be regarded as a matrix of ultrametrics. By \eqref{eqn_reciprocal_clustering}, reciprocal clustering searches for chains that minimize their maximum dissimilarity in the symmetric matrix $\bbarA_X:=\max(A_X, A_X^T)$, where the $\max$ is applied element-wise. This is equivalent to finding chains in $\bbarA_X$ that have minimum cost in a $\ell_\infty$ sense. Likewise, nonreciprocal clustering searches for directed chains of minimum cost in $A_X$ to construct the matrix $\tdu^*_X$ [cf.~\eqref{eqn_nonreciprocal_chains}] and selects the maximum of the directed costs by performing the operation $u^{\NR}_X = \max(\tdu^*_X,\tdu^{*T}_X)$ [cf. \eqref{eqn_nonreciprocal_clustering}]. These operations can be performed algorithmically using matrix powers in the dioid algebra $\mathfrak{A}:= (\reals^+\cup\{+\infty\},\min,\max)$~\cite{GondranMinoux08}.

In $\mathfrak{A}$, the regular sum is replaced by the minimization operator and the regular product by maximization. Indeed, using $\oplus$ and $\otimes$ to denote sum and product, respectively, on this dioid algebra we have $a\oplus b := \min(a,b)$ and $a\otimes b := \max(a,b)$ for all $a, b \in \reals^+\cup\{+\infty\}$. In the algebra $\mathfrak{A}$, the matrix product $A\otimes B$ of two real valued matrices of compatible sizes is therefore given by the matrix with entries
\begin{equation}\label{def_star}
   \big[A \otimes B\big]_{ij}  
       \! := \! \bigoplus_{k=1}^n \big(A_{ik} \otimes B_{kj} \big) 
       \ =\!\! \min_{k\in \{1, .. , n\}} \! \max \big(A_{ik},B_{kj} \big).
\end{equation}

For integers $k\geq 2$ dioid matrix powers $A_X^{k}:=A_X\otimes A_X^{k-1}$ with $A_X^{1}:=A_X$ of a dissimilarity matrix are related to ultrametric matrices $u_X$. We delve into this relationship in the next section.

\subsection{Dioid powers and ultrametrics}
Notice that the elements of the dioid power $u_X^{2}$ of a given ultrametric matrix $u_X$ are given by
\begin{equation}\label{def_diod_algebra_ultrametric}
   \big[u_X^{2}\big]_{ij}  
       = \min_{k\in\{1, .. , n\}} \, \max \big([u_{X}]_{ik},[u_{X}]_{kj} \big).
\end{equation}
Since $u_X$ satisfies the strong triangle inequality we have that $[u_X]_{ij}\leq\max \big([u_{X}]_{ik},[u_{X}]_{kj}\big)$ for all $k\in\{1, .. , n\}$. And for $k=j$ in particular we further have that $\max \big([u_X]_{ij},[u_X]_{jj})=\max\big([u_X]_{ij},0)=[u_X]_{ij}$. Combining these two observations it follows that the result of the minimization in \eqref{def_diod_algebra_ultrametric} is $\big[u_X^{2}\big]_{ij} =  \big[u_X\big]_{ij}$ since none of its arguments is smaller that $[u_{X}]_{ij}$ and one of them is exactly $[u_{X}]_{ij}$. This being valid for all $i,j$ implies 
\begin{equation}\label{def_diod_algebra_ultrametric_result}
   u_X^{2} =  u_X.
\end{equation}
Furthermore, a matrix having the property in \eqref{def_diod_algebra_ultrametric_result} is such that $\big[u_X\big]_{ij} = \big[u_X^{2}\big]_{ij} = \min_{k\in\{1, .. , n\}} \, \max \big([u_{X}]_{ik},[u_{X}]_{kj} \big) \leq \max\big([u_{X}]_{il},[u_{X}]_{lj} \big)$ for all $l$, which is just a restatement of the strong triangle inequality. Therefore, a non-negative matrix $u_X$ represents a finite ultrametric if and only if \eqref{def_diod_algebra_ultrametric_result} is true, has null diagonal elements and positive off-diagonal elements, and is symmetric, $u_X=u_X^T$. From definition \eqref{def_star} it follows that the $l$-th dioid power $A_X^{l}$ is such that its entry $[A_X^{l}]_{ij}$ represents the minimum cost of a chain from node $i$ to $j$ containing at most $l$ hops. We then expect dioid powers to play a key role in the construction of ultrametrics.

The \emph{quasi-inverse} of a matrix in a dioid algebra is a useful concept that simplifies the proofs within this section. In any dioid algebra we call quasi-inverse of $A$, denoted by $A^\dag$, to the limit, when it exists, of the sequence of matrices \cite[Ch.4, Def. 3.1.2]{GondranMinoux08}
\begin{equation}\label{eqn_def_quasi_inverse}
A^\dag := \lim_{k\to \infty } I \oplus A \oplus A^{2} \oplus ... \oplus A^{k},
\end{equation}
where $I$ has zeros in the diagonal and $+ \infty$ in the off-diagonal elements. The utility of the quasi-inverse resides in the fact that, given a dissimilarity matrix $A_X$, then \cite[Ch.6, Sec 6.1]{GondranMinoux08}
\begin{equation}\label{eqn_quasi_inverse_nonrecip}
[A_X^\dag]_{ij} = \min_{C(x_i,x_j)} \,\,\, \max_{k | x_k \in C(x_i,x_j)} \,\, A_X(x_k,x_{k+1}).
\end{equation}
I.e., the elements of the quasi-inverse $A_X^\dag$ correspond to the directed minimum chain costs $\tdu^*_X$ of the associated network $(X, A_X)$ as defined in \eqref{eqn_nonreciprocal_chains}.

\subsection{Algorithms for admissible clustering methods}
The reciprocal and nonreciprocal ultrametrics can be obtained via simple dioid matrix operations, as stated next.

%
\begin{theorem}\label{theo_algo_recip_nonrecip}
For any network $N=(X, A_X)$ with $n$ nodes the reciprocal ultrametric $u^{\R}_X$ defined in \eqref{eqn_reciprocal_clustering} can be computed as
\begin{align}
   u^{\R}_X = \Big(\max\left( {A}_X, A_X^T \right)\Big)^{n-1}, \label{eqn_algo_recip} 
\end{align}
where the matrix operations are in the dioid algebra $\mathfrak{A}$. Similarly, the nonreciprocal ultrametric $u^{\NR}_X$ defined in \eqref{eqn_nonreciprocal_clustering} can be computed as
\begin{align}
   u^{\NR}_X=\max\left({A}_X^{n-1},\left(A_X^T\right)^{n-1}\right). \label{eqn_algo_nonrecip}
\end{align}\end{theorem}
\begin{myproofnoname} 
By comparing \eqref{eqn_quasi_inverse_nonrecip} with \eqref{eqn_nonreciprocal_chains}, we can see that $A_X^\dag=\tdu^*_X$ from where it follows [cf.~\eqref{eqn_nonreciprocal_clustering}]
\begin{align}
u^{\NR}_X = \max \big( A_X^\dag, (A_X^\dag)^T \big) \label{eqn_algo_nonrecip_2}.
\end{align}
Similarly, if we consider the quasi-inverse of the symmetrized matrix $\bbarA_X:=\max(A_X, A_X^T)$, expression \eqref{eqn_quasi_inverse_nonrecip} becomes
\begin{equation}\label{eqn_quasi_inverse_recip}
[\bar{A}_X^\dag]_{ij} = \min_{C(x_i,x_j)} \,\,\, \max_{k | x_k \in C(x_i,x_j)} \,\, \bar{A}_X(x_k,x_{k+1}).
\end{equation}
From comparing \eqref{eqn_quasi_inverse_recip} and \eqref{eqn_reciprocal_clustering} it is immediate that
\begin{equation}\label{eqn_algo_recip_2}
u^{\R}_X= \bar{A}_X^\dag = \big(\!\max(A_X, A_X^T)\big)^\dag.
\end{equation}
If we show that $A^\dag_X=A^{n-1}_X$, then \eqref{eqn_algo_recip_2} and \eqref{eqn_algo_nonrecip_2} imply equations \eqref{eqn_algo_recip} and \eqref{eqn_algo_nonrecip} respectively, completing the proof. 

Notice that in $\mathfrak{A}$, the $\min$ or $\oplus$ operation is idempotent, i.e. $a \oplus a = a$ for all $a$. In this case, it can be shown that \cite[Ch.4, Prop. 3.1.1]{GondranMinoux08}
\begin{equation}\label{eqn_quasi_inverse_develop_0}
I \oplus A_X \oplus A_X^{2} \oplus ... \oplus A_X^{k}= (I \oplus A_X)^{k},
\end{equation}
for all $k \geq 1$. Recalling that $I$ has zeros in the diagonal and $+ \infty$ in the off-diagonal elements, it is immediate that $I \oplus A_X = A_X$. Consequently, \eqref{eqn_quasi_inverse_develop_0} becomes
\begin{equation}\label{eqn_quasi_inverse_develop}
I \oplus A_X \oplus A_X^{2} \oplus ... \oplus A_X^{k}= A_X^{k}.
\end{equation}
Taking the limit to infinity in both sides of equality \eqref{eqn_quasi_inverse_develop} and invoking the definition of the quasi-inverse in \eqref{eqn_def_quasi_inverse}, we obtain
\begin{equation}\label{eqn_quasi_inverse_develop_2}
A_X^\dag= \lim_{k \to \infty} A_X^{k}.
\end{equation}
Finally, it can be shown \cite[Ch. 4, Sec. 3.3, Theo. 1]{GondranMinoux08} that $A_X^{n-1}=A_X^{n}$, proving that the limit in \eqref{eqn_quasi_inverse_develop_2} exists and, more importantly, that $A_X^\dag=A_X^{n-1}$, as desired.
\end{myproofnoname}

For the reciprocal ultrametric we symmetrize dissimilarities with a maximization operation and take the $(n-1)$-th power of the resulting matrix on the dioid algebra $\mathfrak{A}$. For the nonreciprocal ultrametric we revert the order of these two operations. We
first consider matrix powers ${A}_X^{n-1}$ and $\left(A_X^T\right)^{n-1}$ of the dissimilarity matrix and its transpose which we then symmetrize with a maximization operator. Besides emphasizing the extremal nature (cf.~Theorem~\ref{theo_extremal_ultrametrics}) of reciprocal and nonreciprocal clustering, Theorem \ref{theo_algo_recip_nonrecip} suggests the existence of intermediate methods in which we raise dissimilarity matrices $A_X$ and $A_X^T$ to some power, perform a symmetrization, and then continue applying matrix powers. These procedures yield methods that are not only valid but coincide with the family of semi-reciprocal ultrametrics introduced in Section \ref{sec_inter_reciprocal}, as the following proposition asserts.

\begin{proposition}\label{prop_general_algo} For any network $N=(X, A_X)$ with $n$ nodes the $t$-th semi-reciprocal ultrametric $u_X^{\SR(t)}$ in \eqref{eqn_inter_reciprocal_clustering} for every natural $t \geq 2$ can be computed as
\begin{equation}\label{eqn_algo_semi_reciprocal_2}
   u_X^{\SR(t)} = \left(\max\left({A}_X^{t-1},\left(A_X^T\right)^{t-1}\right)\right)^{n-1},
\end{equation}
where the matrix operations are in the dioid algebra $\mathfrak{A}$.
\end{proposition}
\begin{myproofnoname}
By comparison with \eqref{eqn_algo_recip}, in \eqref{eqn_algo_semi_reciprocal_2} we in fact compute reciprocal clustering on the network $(X, A^{t-1}_X)$. Furthermore, from the definition of matrix multiplication \eqref{def_star} in $\mathfrak{A}$, the $(t-1)$-th dioid power $A_X^{t-1}$ is such that its entry $[A_X^{t-1}]_{ij}$ represents the minimum cost of a chain containing at most $t$ nodes, i.e.
\begin{equation}\label{eqn_dioid_semi_recip}
[A_X^{t-1}]_{ij} = \min_{C_t(x_i, x_j)} \,\,\,  \max_{k | x_k\in C_t(x_i, x_j)} A_X(x_k, x_{k+1}).
\end{equation} 
It is just a matter of notation, when comparing \eqref{eqn_dioid_semi_recip} and \eqref{eqn_inter_cost} to see that $A_X^{t-1} = A^{\SR(t)}_X$. Since semi-reciprocal clustering is equivalent to applying reciprocal clustering to network $(X, A^{\SR(t)}_X)$ [cf. \eqref{eqn_inter_reciprocal_clustering} and \eqref{eqn_reciprocal_clustering}], the proof concludes.
\end{myproofnoname}

The result in \eqref{eqn_algo_semi_reciprocal_2} is intuitively clear. The powers ${A}_X^{t-1}$ and $\left(A_X^T\right)^{t-1}$ represent the minimum cost among directed chains of at most $t-1$ links. In the terminology of Section \ref{sec_inter_reciprocal} these are the costs of optimal secondary chains containing at most $t$ nodes. Therefore, the maximization $\max\big({A}_X^{t-1},\left(A_X^T\right)^{t-1}\big)$ computes the cost of joining two  nodes with secondary chains of at most $t$ nodes in each direction. This is the definition of $\bar{A}^{\SR(t)}_X$ in \eqref{eqn_inter_reciprocal_clustering}. Applying the $(n-1)$-th dioid power to this new matrix is equivalent to looking for minimizing chains in the network with costs given by the secondary chains. Thus, the outermost dioid power computes the costs of the optimal main chains that achieve the ultrametric values in \eqref{eqn_inter_reciprocal_clustering}. 

Observe that we recover \eqref{eqn_algo_recip} by making $t=2$ in \eqref{eqn_algo_semi_reciprocal_2} and that we recover \eqref{eqn_algo_nonrecip} when $t=n$. For this latter case note that when $t=n$ in \eqref{eqn_algo_semi_reciprocal_2}, comparison with \eqref{eqn_algo_nonrecip} shows that $\max({A}_X^{t-1},(A_X^T)^{t-1})=\max({A}_X^{n-1},(A_X^T)^{n-1})=u^{\NR}_X$. However, since $u^{\NR}_X$ is an ultrametric it is idempotent in the dioid algebra [cf. \eqref{def_diod_algebra_ultrametric_result}] and the outermost dioid power in \eqref{eqn_algo_semi_reciprocal_2} is moot. This recovery is consistent with the observations in \eqref{eqn_inter_reciprocal_clustering_1} and \eqref{eqn_inter_reciprocal_clustering_2} that reciprocal and nonreciprocal clustering are particular cases of semi-reciprocal clustering $\ccalH^{\SR(t)}$ such that for $t=2$ we have $u^{\SR(2)}_X=u^{\R}_X$ and for $t\geq n$ it holds that $u^{\SR(t)}_X=u^{\NR}_X$. The results in Theorem \ref{theo_algo_recip_nonrecip} and Proposition \ref{prop_general_algo} emphasize the extremal nature of the reciprocal and nonreciprocal methods and characterize the semi-reciprocal ultrametrics as natural intermediate clustering methods in an algorithmic sense.

This algorithmic perspective allows for a generalization in which the powers of the matrices $A_X$ and $A_X^T$ are different. To be precise consider positive integers $t,t'>0$ and define the algorithmic intermediate clustering method $\ccalH^{t,t'}$ with parameters $t,t'$ as the one that maps the given network $N=(X, A_X)$ to the ultrametric set $(X,u^{t,t'}_X) = \ccalH^{t,t'}(N)$ given by
\begin{equation}\label{eqn_algo_algorithmic_intermediate}
   u^{t,t'}_X := \left(\max\left({A}_X^{t},\left(A_X^T\right)^{t'}\right)\right)^{n-1}. 
\end{equation}
The ultrametric \eqref{eqn_algo_algorithmic_intermediate} can be interpreted as a semi-reciprocal ultrametric where the allowed length of secondary chains varies with the direction. Forward secondary chains may have at most $t+1$ nodes whereas backward secondary chains may have at most $t'+1$ nodes. The algorithmic intermediate family $\ccalH^{t,t'}$ encapsulates the semi-reciprocal family since $\ccalH^{t,t}\equiv\ccalH^{\SR(t+1)}$ as well as the reciprocal method since $\ccalH^{\R}\equiv\ccalH^{1,1}$ as it follows from comparison of \eqref{eqn_algo_algorithmic_intermediate} with \eqref{eqn_algo_semi_reciprocal_2} and \eqref{eqn_algo_recip}, respectively. We also have that $\ccalH^{\NR}(N) = \ccalH^{n-1, n-1}(N)$ for all networks $N=(X, A_X)$ such that $|X| \leq n$. This follows from the comparison of \eqref{eqn_algo_algorithmic_intermediate} with \eqref{eqn_algo_nonrecip} and the idempotency of $u^{\NR}_X=\max({A}_X^{n-1},(A_X^T)^{n-1})$ with respect to the dioid algebra. The intermediate algorithmic methods $\ccalH^{t,t'}$ are admissible as we claim in the following proposition.

\begin{proposition}\label{prop_algorithmic_intermediate_ultrametric} The hierarchical clustering method $\ccalH^{t,t'}$ is valid and admissible. I.e., $u_X^{t,t'}$ defined in \eqref{eqn_algo_algorithmic_intermediate} is a valid ultrametric and $\ccalH^{t,t'}$ satisfies axioms (A1)-(A2).
\end{proposition}
\begin{myproofnoname} 
Since method $\ccalH^{t,t'}$ is a generalization of $\ccalH^{\SR(t)}$, the proof is almost identical to the one of Proposition \ref{inter_reciprocal_axioms}. The only major difference is that showing symmetry of $u^{t,t'}_X$, i.e. $u^{t,t'}_X(x, x')=u^{t,t'}_X(x', x)$ for all $x, x' \in X$, is not immediate as in the case of $u_X^{\SR(t)}$.
In a fashion similar to \eqref{eqn_inter_reciprocal_clustering}, we rewrite the definition of $u^{t,t'}_X$ given an arbitrary network $(X, A_X)$ in terms of minimizing chains,
\begin{equation}\label{eqn_proof_algo_general_1}
u^{t,t'}_X(x,x')= \min_{C(x,x')} \,\,\, \max_{i | x_i\in C(x,x')} A^{t,t'}_X(x_i, x_{i+1})
\end{equation}
where the function $A^{t,t'}_X$ is defined as
\begin{equation}\label{eqn_proof_algo_general_2}
A^{t,t'}_X\!(x,x'):= \max\! \left(\!A^{\SR(t+1)}_X\!(x,x'), A^{\SR(t'+1)}_X\!(x',x)\!\right),
\end{equation}
for all $x, x' \in X$ and functions $A^{\SR(\cdot)}_X$ as defined in \eqref{eqn_inter_cost}. 
Notice that $A^{t,t'}_X$ is not symmetric in general. Symmetry of $u^{t,t'}_X$, however, follows from the following claim.

\begin{claim}\label{claim_algorithmic_intermediate}
Given any network $(X, A_X)$ and a pair of nodes $x, x' \in X$ such that $u^{t,t'}_X(x, x') = \delta$, then $u^{t,t'}_X(x', x) \leq \delta$.
\end{claim}
\begin{myproofnoname}
Assuming $u^{t,t'}_X(x, x')=\delta$, we denote by $C(x, x')=[x=x_0, x_1, ... , x_l=x']$ a minimizing main chain achieving the cost $\delta$ in \eqref{eqn_proof_algo_general_1}. 
Thus, we must show that there exists a main chain $\hat{C}(x', x)$ from $x'$ back to $x$ with cost not exceeding $\delta$. From definition \eqref{eqn_proof_algo_general_2}, there must exist secondary chains in both directions between every pair of consecutive nodes $x_i, x_{i+1}$ in $C(x, x')$ with cost no greater than $\delta$. These secondary chains $C_{t+1}(x_i, x_{i+1})$ and $C_{t'+1}(x_{i+1}, x_{i})$ can have at most $t+1$ nodes in the forward direction and at most $t'+1$ nodes in the opposite direction. Moreover, without loss of generality we may consider the secondary chains as having exactly $t+1$ nodes in one direction and $t'+1$ in the other if we do not require consecutive nodes to be distinct. 

Focus on a pair of consecutive nodes $x_i, x_{i+1}$ of the main chain $C(x, x')$. If we can construct a main chain from $x_{i+1}$ back to $x_i$ with cost not greater than $\delta$, then we can concatenate these chains for pairs $x_{i+1}, x_i$ for all $i$ and obtain the required chain $\hat{C}(x', x)$ in the opposite direction.

Notice that the secondary chains $C_{t'+1}(x_{i+1}, x_{i})$ and $C_{t+1}(x_i, x_{i+1})$ can be concatenated to form a loop $L(x_{i+1},x_{i+1})$, i.e. a chain starting and ending at the same node, of $t'+t+1$ nodes and cost not larger than $\delta$. We rename the nodes in $L(x_{i+1},x_{i+1})=[x_{i+1}=x^{0}, x^{1}, ... , x^{t'}=x_i, ..., x^{t'+t-1}, x^{t'+t}=x_{i+1}]$ starting at $x_{i+1}$ and following the direction of the loop.

Now we are going to construct a main chain $C(x_{i+1}, x_i)$ from $x_{i+1}$ to $x_i$. We may reinterpret the loop $L(x_{i+1},x_{i+1})$ as the concatenation of two secondary chains $[x^{0}, x^1, \ldots, x^{t}]$ and $[x^t, x^{t+1}, \ldots, x^{t+t'}=x^0]$ each of them having cost not greater than $\delta$. Thus, we may pick $x^{0}=x_{i+1}$ and $x^{t}$ as the first two nodes of the main chain $C(x_{i+1}, x_i)$. With the same reasoning, we may link $x^{t}$ with $x^{\,2t \!\!\! \mod \!(t+t')}$ with cost not exceeding $\delta$, and we may link $x^{\,2t \!\!\! \mod \!(t+t')}$ with $x^{\,3t \!\!\! \mod \!(t+t')}$ with cost not exceeding $\delta$, and so on. Hence, we construct the main chain
\begin{align}\label{eqn_proof_algo_general_3}
C(x_{i+1}, x_i) \! = \! [x^0, x^t, x^{2t \!\!\! \mod \!\!(t+t')}, \ldots , x^{(t+t'-1)t \!\!\! \mod\! \!(t+t')}], 
\end{align}
which, by construction, has cost not exceeding $\delta$.

In order to finish the proof, we need to verify that the last node in the chain in \eqref{eqn_proof_algo_general_3} is in fact $x^{t'} = x_i$. To do so, we have to show that $(t+t'-1) \, t \equiv t' \mod (t+t')$, which follows from rewriting the left-hand side as $(t+t') (t-1) + t'$.
\end{myproofnoname}

Applying Claim \ref{claim_algorithmic_intermediate} to an arbitrary pair of nodes $x, x'$ and then to the pair $x', x$ implies that $u^{t,t'}_X(x, x') = u^{t,t'}_X(x', x)$, as needed to show Proposition~\ref{prop_algorithmic_intermediate_ultrametric}.
\end{myproofnoname}

Algorithms to compute ultrametrics associated with the grafting families in Section \ref{sec_grafting} entail simple combinations of matrices $u^{\R}_X$ and $u^{\NR}_X$. E.g., the ultrametrics in \eqref{def_mu_beta_1} corresponding to the grafting method $\ccalH^{\R/\NR}(\beta)$ can be computed as
\begin{equation}\label{eqn_algo_algorithmic_grafting}
   u^{\R/\NR}_X(\beta) =   u_X^{\NR}\circ\ind{u_X^{\R}\leq\beta}  
                       + u_X^{\R} \circ\ind{u_X^{\R}>\beta},
\end{equation}
where $\circ$ denotes the Hadamard matrix product and $\ind{\cdot}$ is an element-wise indicator function.

In symmetric networks, Theorem~\ref{theo_extremal_ultrametrics} states that any admissible method must output an ultrametric equal to the single linkage ultrametric, that we can denote by $u^{\SL}_X$. Thus, all algorithms in this section yield the same output $u^{\SL}_X$ when restricted to symmetric matrices $A_X$. Considering, e.g., the algorithm for the reciprocal ultrametric in \eqref{eqn_algo_recip} and noting that for a symmetric network $A_X=\max(A_X, A^T_X)$ we conclude that single linkage can be computed as
\begin{equation}\label{eqn_algorithm_single_linkage}
u^{\SL}_X = A_X^{n-1}.
\end{equation}
Algorithms for the convex combination family in Section \ref{sec_convex_comb} involve computing dioid algebra powers of a convex combination of ultrametric matrices. Given two admissible methods $\ccalH^1$ and $\ccalH^2$ with outputs $(X, u^1_X)=\ccalH^1(N)$ and $(X, u^2_X)=\ccalH^2(N)$, and  $\theta\in[0,1]$, the ultrametric in \eqref{eqn_def_u_1_2_bar} corresponding to the method $\ccalH^{12}_\theta$ can be computed as
\begin{equation}\label{eqn_algo_convex_comb}
u^{12}_X(\theta)= \Big(\theta\, u^1_X + (1-\theta) \, u^2_X\Big)^{n-1}.
\end{equation}
The operation $\theta\, u^1_X + (1-\theta) \, u^2_X$ is just the regular convex combination in 
\eqref{eqn_def_sym_net_conv_comb} and the dioid power in \eqref{eqn_algo_convex_comb} implements the single linkage operation in \eqref{eqn_def_u_1_2_bar} as it follows from \eqref{eqn_algorithm_single_linkage}.

\begin{remark}\normalfont 
It follows from \eqref{eqn_algo_recip}, \eqref{eqn_algo_nonrecip}, \eqref{eqn_algo_semi_reciprocal_2}, \eqref{eqn_algo_algorithmic_intermediate}, \eqref{eqn_algo_algorithmic_grafting}, and \eqref{eqn_algo_convex_comb} that all methods presented in this paper can be computed in a number of operations of order $O(n^4)$ which coincides with the time it takes to compute $n$ matrix products of matrices of size $n\times n$. This complexity can be reduced to $O(n^3\log n)$ by noting that the dioid matrix power $A^n$ can be computed via the sequence $A, A^2, A^4, \ldots$ which requires $O(\log n)$ matrix products at a cost of $O(n^3)$ each. Complexity can be further reduced using the sub cubic dioid matrix multiplication algorithms in \cite{VassilevskaEtal09, DuanPettie09} that have complexity $O(n^{2.688})$ for a total complexity of $O(n^{2.688}\log n)$ to compute the $n$-th matrix power. There are also related methods with even lower complexity. For the case of reciprocal clustering, complexity of order $O(n^2)$ can be achieved by leveraging an equivalence between single linkage and a minimum spanning tree problem \cite{Hu61,daniel}. For the case of nonreciprocal clustering, Tarjan's method \cite{tarjan-improved} can be implemented to reduce complexity to $O(n^2\log n)$. 
\end{remark}

\section{Numerical results}\label{sec_numerical_results}

The U.S. Department of Commerce publishes a yearly table of input and outputs organized by economic sectors\footnote{Available at http://www.bea.gov/industry/io\_annual.htm}. We focus on a particular section of this table, called \emph{uses}, that corresponds to the inputs to production for year 2011. More precisely, we are given a set $I$ of 61 industrial sectors as defined by the North American Industry Classification System and a similarity function $U\!:\! I \! \times \! I \to \reals_+$ where $U(i, i')$ represents how much of the production of sector $i$, expressed in dollars, is used as an input of sector $i'$. Based on this, we define the network $N_I=(I, A_I)$ where the dissimilarity function $A_I$ satisfies $A_I(i,i)=0$ for all $i\in I$ and, for $i \neq i' \in I$, is given by
\begin{equation}\label{eqn_def_io_dissimilarity}
A_I(i, i') := 1- \frac{U(i, i')}{\sum_j U(j, i')}.
\end{equation}
The normalization $U(i, i')/\sum_j U(j, i')$ in \eqref{eqn_def_io_dissimilarity} can be interpreted as the proportion of the input in dollars to productive sector $i'$ that comes from sector $i$. In this way, we focus on the combination of inputs of a sector rather than the size of the economic sector itself. That is, a small dissimilarity from sector $i$ to sector $i'$ implies that sector $i'$ highly relies on the output of sector $i$ as input for its own production. 

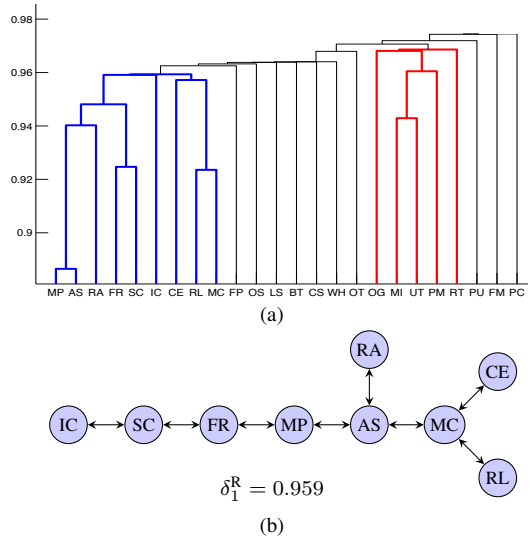
\begin{figure}
\centering
\centerline{\input{figures/reciprocal_example_io_small.tex} }
\vspace{-0.1in}
\caption{(a) Partial view of the reciprocal dendrogram output by $\ccalH^\R$ when applied to $N_I$. Two clusters formed at resolutions $\delta^\R_1=0.959$ and $\delta^\R_2=0.969$ are highlighted in blue and red, respectively. (b)~Detail of blue cluster. Edges represent bidirectional influence between adjacent sectors.}
\vspace{-0.2in}
\label{fig_reciprocal_example_io}
\end{figure}

\vspace{0.05in}\noindent\textbf{Reciprocal clustering.}
The outcome of applying the reciprocal clustering method $\ccalH^\R$ defined in \eqref{eqn_reciprocal_clustering} to the network $N_I$ is computed with the formula in \eqref{eqn_algo_recip}. A partial view of the resulting dendrogram is shown in Fig. \ref{fig_reciprocal_example_io}-(a) where two clusters appearing at resolutions $\delta^\R_1=0.959$ and $\delta^\R_2=0.969$ are highlighted in blue and red, respectively. We also depict in Fig. \ref{fig_reciprocal_example_io}-(b) the nodes in the blue cluster with edges representing bidirectional influence between industrial sectors at the corresponding resolution. That is, a double arrow is drawn between two nodes if and only if the dissimilarity between these nodes in both directions is less than or equal to $\delta^\R_1$. In particular, it shows the bidirectional chains of minimum cost between two nodes. E.g., the bidirectional chain of minimum cost from the sector `Rental and leasing services of intangible assets' (RL) to `Computer and electronic products' (CE) goes through `Management of companies and enterprises' (MC). 

%

It follows from \eqref{eqn_reciprocal_clustering} that the reciprocal clustering method $\ccalH^\R$ tends to cluster sectors of balanced influence in both directions. E.g., the first two sectors to be merged by $\ccalH^\R$ are `Administrative and support services' (AS) and `Miscellaneous professional, scientific and technical services' (MP) at resolution $\delta=0.887$. This occurs because 13.2\% of the input of AS comes from MP -- corresponding to $A_I(\text{MP}, \text{AS})=0.868$ -- and 11.3\% of MP's input comes from AS, both influences being similar in magnitude. It is reasonable that these two sectors hire services from each other in order to better deliver their own services. This balanced behavior is more frequent among service sectors than between raw material extraction (primary) or manufacturing (secondary) sectors. Indeed, the blue cluster in Fig. \ref{fig_reciprocal_example_io}-(b) is mainly composed of services. The first two mergings occur between MP-AS and RL-MC representing professional, support, rental and management services, respectively. At resolution $\delta=0.925$, the sectors `Federal Reserve banks, credit intermediation, and related activities' (FR) and `Securities, commodity contracts, and investments' (SC) merge. This is an exception to the described balanced mergings between service sectors. Indeed, 24.1\% of FR's input comes from SC whereas only 7.5\% of SC's input comes from FR. This is expected since credit intermediation entities in FR have as input investments done in the SC sector. At resolution $\delta=0.940$, `Real estate' (RA) joins the MP-AS cluster due to a bidirectional influence between RA and AS. More precisely, 6.5\% of the input to the RA sector comes from AS and 6.0\% vice versa. This implies that the RA sector hires external administrative and support services and the AS sector depends on the real estate services to, e.g., rent locations for their operation.
The MP-AS-RA cluster merges with the FR-SC cluster at resolution $\delta=0.948$ due to the relation between MP and FR. Indeed, MP provides 11.3\% of FR input -- corresponding to $A_I(\text{MP}, \text{FR})=0.887$ -- and 5.2\% of MP's input comes from FR. At resolution $\delta=0.957$, CE joins the RL-MC cluster due to its bidirectional influence relation with MC. The sector of electronic products CE is the only sector in the blue cluster formed at resolution $\delta^\R_1=0.959$ that does not represent a service. The `Insurance carriers and related activities' (IC) sector joins the MP-AS-RA-FR-SC cluster at resolution $\delta=0.959$ because of its relation with SC. In fact, 4.5\% of IC's input comes from SC in the form of securities and investments and 4.1\% of SC's input comes from IC in the form of insurance policies for investments. Finally, at resolution $\delta^\R_1=0.959$, the clusters MP-AS-RA-FR-SC-IC and CE-RL-MC merge due to the relation between the supporting services AS and the management services MC.

Requiring direct bidirectional influence generates some clusters which are counter-intuitive. E.g., in the reciprocal dendrogram in Fig. \ref{fig_reciprocal_example_io}-(a), at resolution $\delta=0.971$ when the blue and red clusters merge together we have that the `Oil and gas extraction' sector (OG) in the red cluster joins, e.g., the insurance sector IC in the blue cluster. However, OG does not merge with `Petroleum and coal products' (PC), a sector that one would expect to be more closely related, until resolution $\delta=0.975$. In order to avoid this situation, we may allow nonreciprocal influence as we do next.

\begin{figure}
\centering
\centerline{\input{figures/nonreciprocal_example_io_small.tex} }
\vspace{-0.1in}
\caption{(a) Partial view of the nonreciprocal dendrogram output by $\ccalH^{\NR}$ when applied to $N_I$. One cluster, formed at resolution $\delta^{\NR}_1=0.900$, is highlighted in blue. (b) Detail of highlighted cluster. Directed edges between sectors imply unidirectional influence between them. Thick arrows mark the longest cycle.}
\vspace{-0.15in}
\label{fig_nonreciprocal_example_io}
\end{figure}
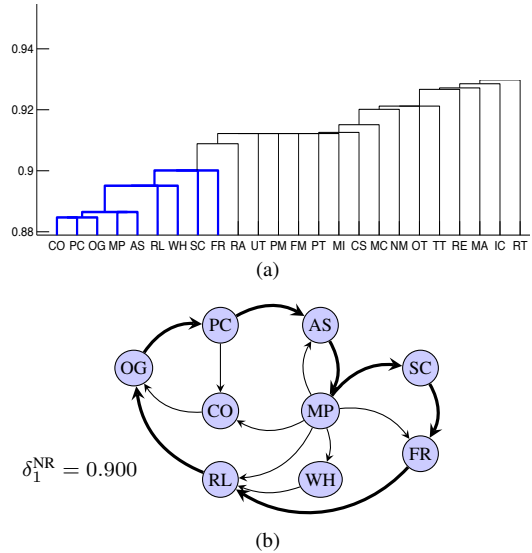

\vspace{0.05in}\noindent\textbf{Nonreciprocal clustering.}
The outcome of applying the nonreciprocal clustering method $\ccalH^\NR$ defined in \eqref{eqn_nonreciprocal_clustering} to $N_I$ is computed via \eqref{eqn_algo_nonrecip}. A partial view of the resulting dendrogram is shown in Fig. \ref{fig_nonreciprocal_example_io}-(a). Let us first observe that the nonreciprocal ultrametrics (merging resolutions) in Fig. \ref{fig_nonreciprocal_example_io}-(a) are not larger than the reciprocal ultrametrics in Fig. \ref{fig_reciprocal_example_io}-(a), as it should be the case given the inequality in \eqref{eqn_theo_extremal_ultrametrics}. As a test case we have that the `Mining, except oil and gas' (MI) and the `Utilities' (UT) sectors become part of the same cluster in the reciprocal dendrogram at a resolution $\delta=0.943$ whereas they merge in the nonreciprocal dendrogram at resolution $\delta'=0.912<0.943$.

A more interesting observation is that the nonreciprocal dendrogram is qualitatively very different from the reciprocal dendrogram. In the reciprocal dendrogram we tended to see the formation of definite clusters that then merged into larger clusters at coarser resolutions. In the nonreciprocal dendrogram, in contrast, we see the progressive agglutination of economic sectors into a central cluster. Indeed, the first non-singleton cluster to arise is formed at resolution $\delta =0.885$ by the sectors of oil and gas extraction OG, petroleum and coal products PC, and `Construction' (CO). In Fig.~\ref{fig_nonreciprocal_example_io}-(b) we see that this cluster forms due to the influence cycle $[$OG, PC, CO, OG$]$. Of all the economic input to PC, $82.6\%$ comes from the OG sector -- which is represented by the dissimilarity $A_I(\text{OG}, \text{PC})=0.174$ -- in the form of raw material for its productive processes of which oil refining is the dominant one. In the input to CO a total of $11.5\%$ comes from PC as fuel and lubricating oil for heavy machinery as well as asphalt coating, and $12.3\%$ of OG's input comes from CO mainly from engineering projects to enable extraction such as perforation and the construction of pipelines and their maintenance.

At resolution $\delta =0.887$ this cluster grows by the simultaneous incorporation of the support service sector AS and the professional service sector MP. These sectors join due to the loop $[$AS, MP, CO, OG, PC, AS$]$. The three new edges in this loop that involve the new sectors are the ones from PC to AS, from AS to MP and from MP to CO. Of all the economic input to AS, $13.4\%$ comes from the PC sector in the form of, e.g., fuel for the transportation of manpower. Of MP's input, 11.3\% comes from AS corresponding to administrative and support services hired by the MP sector for the correct delivery of MP's professional services and in the input to CO a total of 12.8\% comes from MP from, e.g., architecture and consulting services for the construction. 
We then see the incorporation of the rental service sector RL and `Wholesale trade' (WH) to the five-node cluster at resolution $\delta =0.895$ given by the loop $[$WH, RL, OG, PC, AS, MP, WH$]$. Finally, at resolution $\delta^{\NR}_1=0.900$ the financial sectors SC and FR join this cluster due to the chain $[$SC, FR, RL, OG, PC, AS, MP, SC$]$.

The nonreciprocal clustering method $\ccalH^{\NR}$ detects cyclic influences which, in general, lead to clusters that are more reasonable than those requiring bidirectional influence as in reciprocal clustering. E.g., $\ccalH^{\NR}$ merges the oil and gas OG and petroleum products PC sectors at resolution $\delta=0.885$ before they merge with the insurance sector IC at resolution $\delta=0.923$. By contrast, as has been already stated, $\ccalH^{\R}$ merges OG with IC before their common joining with PC. However, the preponderance of cyclic influences in the network of economic interactions $N_I$ leads to the formation of clusters that look more like artifacts than fundamental features. E.g., the cluster that forms at resolution $\delta =0.887$ has AS and MP joining the three-node cluster CO-PC-OG because of an influence cycle of five nodes. From our discussion above, it is thus apparent that allowing clusters to be formed by arbitrarily long cycles overlooks important bidirectional influences between co-clustered nodes. If we wanted a clustering method which at resolution $\delta =0.887$ would cluster the nodes PC, CO, and OG into one cluster and AS and MP into another cluster, we should allow influence to propagate through cycles of at most three or four nodes. A family of methods that permits this degree of flexibility is the family of semi-reciprocal methods $\ccalH^{\SR(t)}$ that we discussed in Section \ref{sec_inter_reciprocal} and whose application we exemplify next.

%
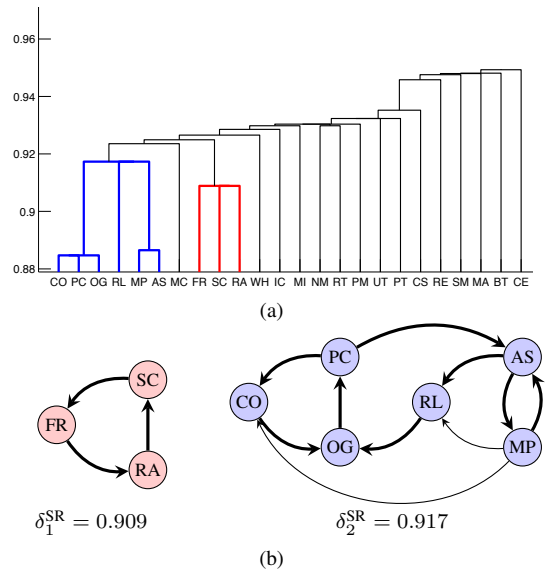
\begin{figure}
\centering
\centerline{\input{figures/semi_reciprocal_example_io_small.tex} }
\vspace{-0.1in}
\caption{(a) Partial view of the semi-reciprocal dendrogram output by $\ccalH^{\SR(3)}$ when applied to $N_I$. Two clusters formed at resolutions $\delta^\SR_1=0.909$ and $\delta^\SR_2=0.917$ are highlighted in red and blue, respectively. (b) Detail of highlighted clusters. Cyclic influences can be observed.}
\vspace{-0.1in}
\label{fig_semi_reciprocal_example_io}
\end{figure}

\vspace{0.05in}\noindent\textbf{Semi-reciprocal clustering.}
The outcome of applying the semi-reciprocal clustering method $\ccalH^{\SR(3)}$ defined in Section \ref{sec_inter_reciprocal} to $N_I$ is computed with the formula in \eqref{eqn_algo_semi_reciprocal_2}. A partial view of the resulting dendrogram is shown in Fig. \ref{fig_semi_reciprocal_example_io}-(a). Two clusters generated at resolutions $\delta^\SR_1=0.909$ and $\delta^\SR_2=0.917$ are highlighted in red and blue, respectively. These clusters are depicted in Fig. \ref{fig_semi_reciprocal_example_io}-(b) with directed edges between the nodes representing dissimilarities less than or equal to the corresponding resolution. E.g., for the cluster generated at resolution $\delta^\SR_1=0.909$, we draw an edge from sector $i$ to sector $i'$ if and only if $A_I(i, i') \leq \delta^\SR_1$. 
Comparing the semi-reciprocal dendrogram in Fig. \ref{fig_semi_reciprocal_example_io}-(a) with the reciprocal and nonreciprocal dendrograms in Figs. \ref{fig_reciprocal_example_io}-(a) and \ref{fig_nonreciprocal_example_io}-(a), we observe that semi-reciprocal clustering merges any pair of sectors at a resolution not higher than the resolution at which they are co-clustered by reciprocal clustering and not lower than the one at which they are co-clustered by nonreciprocal clustering. E.g., the financial sectors FR and SC become part of the same cluster at resolutions $\delta_\R=0.925$ in the reciprocal dendrogram, $\delta_{\SR}=0.909$ in the semi-reciprocal dendrogram and $\delta_{\NR}=0.900$ in the nonreciprocal dendrogram, satisfying $\delta_{\NR} \leq \delta_{\SR} \leq \delta_\R$. This ordering of the merging resolutions is as it should be since the reciprocal and nonreciprocal ultrametrics uniformly bound the output ultrametric of any admissible clustering method such as semi-reciprocal clustering [cf. \eqref{eqn_theo_extremal_ultrametrics}].

The clustering method $\ccalH^{\SR(3)}$ allows reasonable cyclic influences and is insensitive to intricate influences described by long cycles. As already mentioned, $\ccalH^\R$ does not recognize the obvious relation between the oil and gas OG and the petroleum products PC sectors because it requires direct bidirectional influence whereas $\ccalH^{\NR}$ merges OG and PC at a low resolution but also considers other counter-intuitive cyclic influence structures represented by long loops such as the merging of the service sectors AS and MP with the cluster OG-PC-CO before forming a cluster by themselves [cf. Fig. \ref{fig_nonreciprocal_example_io}]. The semi-reciprocal method $\ccalH^{\SR(3)}$ combines the desirable features of reciprocal and nonreciprocal clustering. Indeed, as can be seen from Fig. \ref{fig_semi_reciprocal_example_io}-(a), $\ccalH^{\SR(3)}$ recognizes the heavy industry cluster OG-PC-CO since these three sectors are the first to merge at resolution $\delta=0.885$. However, the service sectors MP and AS form a cluster of their own before merging with the heavy industry cluster. To be more precise, MP and AS merge at resolution $\delta=0.887$ due to the bidirectional influence between them. 
When we increase the resolution, at $\delta^{\SR}_2=0.917$ the `Rental and leasing services' (RL) sector acts as an intermediary merging the OG-PC-CO cluster with the MP-AS cluster forming the blue cluster in Fig. \ref{fig_semi_reciprocal_example_io}-(b). The cycle containing RL with secondary chains of length at most 3 nodes is $[$RL, OG, PC, AS, RL$]$. The sector RL uses administrative and support services from AS to provide their own leasing services, and leasing is a common practice in the OG sector. Thus, we obtain the influences depicted in the blue cluster. At resolution $\delta^{\SR}_1=0.909$ the credit intermediation sector FR, the investment sector SC and the real estate sector RA form a three-node cluster given by the influence cycle $[$RA, SC, FR, RA$]$ and depicted in red in Fig. \ref{fig_semi_reciprocal_example_io}-(b). Of all the economic input to SC, $9.1\%$ comes from the RA sector in the form of, e.g., leasing services related to real estate investment trusts. The sector SC provides 24.1\% of FR's input whereas FR represents 35.1\% of RA's input. 
Notice that in the nonreciprocal dendrogram in Fig. \ref{fig_nonreciprocal_example_io}-(a), these three sectors join the main blue cluster separately due to the formation of intricate influence loops. The semi-reciprocal method, by not allowing the formation of long loops, distinguishes the more reasonable cluster formed by FR-RA-SC.

\section{Conclusion}\label{sec_conclusion}

We identified and described three families of hierarchical clustering methods that, by satisfying the axioms of value and transformation, are contained between reciprocal and nonreciprocal clustering: i) The grafting methods are defined by exchanging branches between the reciprocal and nonreciprocal dendrograms; ii) The convex combination methods are built around the definition of a convex operation in the space of dendrograms; and iii) The semi-reciprocal clustering methods allow the generation of clusters via cyclic influence of a fixed maximum length. Algorithms for the application of the methods described throughout the paper were developed via matrix operations in a min-max dioid algebra. The reciprocal ultrametric was computed by first symmetrizing directed dissimilarities to their maximum and then computing increasing powers of the symmetrized dissimilarity matrix until stabilization whereas, for the nonreciprocal case, the opposite was shown to be true. In a similar fashion, algorithms for the remaining clustering methods presented throughout the paper were developed in terms of finite matrix powers, thus exhibiting computational tractability of our clustering constructions. Finally, we applied the derived clustering methods and algorithms to study the relationship between economic sectors in the United States. As a future research avenue, we seek to further winnow the set of admissible methods by requiring additional properties such as stability -- when clustering similar networks we should obtain similar dendrograms -- and scale invariance -- the formation of clusters should not depend on the scale used to measure dissimilarities.

\bibliographystyle{unsrt}
\bibliography{clustering_biblio}
\end{document}

%% file: figures/tikz_styles.tex
\tikzstyle{phantom vertex} = [ ellipse, 
                               anchor = center, 
                               minimum height = 1*\unit, 
                               minimum width  = 1*\unit,
                               inner sep=0pt,
                               anchor=center]
\tikzstyle{red vertex}   = [black, fill = red!20,   phantom vertex, draw]
\tikzstyle{black vertex} = [black, fill = black!20, phantom vertex, draw]
\tikzstyle{blue vertex}  = [black, fill = blue!20,  phantom vertex, draw]
\tikzstyle{green vertex} = [black, fill = green!20,  phantom vertex, draw]
\tikzstyle{vertex}       = [draw, phantom vertex]

\tikzstyle{point} = [ellipse, inner sep=0pt, draw, fill=white, anchor = center,
                     minimum height = 0.05*\unit, minimum width  = 0.05*\unit]

%% file: my_sections.tex
\usepackage{needspace}

\newcommand{\myparagraph}[1]{\needspace{1\baselineskip}\medskip\noindent {\it #1.}}



%% file: figures/reciprocal_path.tex
\def \thisplotscale {0.5}
\def \unit {\thisplotscale cm}

{\small
\begin{tikzpicture}[-stealth, shorten >=2 ,scale = \thisplotscale, font=\footnotesize]

	\node[blue vertex] (x) {$x$} ;

	\path (x) ++ (4,0)   node [blue vertex]    (x1)   {$x_1$} 
	          ++ (4,0)   node [phantom vertex] (x2)   {$\ldots$}
	          ++ (0.5,0) node [phantom vertex] (xlm1) {$\ldots$} 
	          ++ (4,0)   node [blue vertex]    (xl)   {$x_{l-1}$}
	          ++ (4,0)    node [blue vertex]   (xp)   {$x'$}; 

	\path (x)    edge [bend left, above] node {$A_X(x,x_1)$}       (x1);	
	\path (x1)   edge [bend left, above] node {$A_X(x_1,x_2)$}     (x2);	
	\path (xlm1) edge [bend left, above] node {$A_X(x_{l-2},x_{l-1})$} (xl);	
	\path (xl)   edge [bend left, above] node {$A_X(x_{l-1},x')$}      (xp);	

	\path (x1)   edge [bend left, below] node {$A_X(x_1,x)$}       (x);
	\path (x2)   edge [bend left, below] node {$A_X(x_2,x_1)$}     (x1);
	\path (xl)   edge [bend left, below] node {$A_X(x_{l-1},x_{l-2})$} (xlm1);
	\path (xp)   edge [bend left, below] node {$A_X(x',x_{l-1})$}      (xl);

\end{tikzpicture}
}

%% file: figures/nonreciprocal_path.tex
\def \thisplotscale {0.5}
\def \unit {\thisplotscale cm}
{\small
\begin{tikzpicture}[-stealth, shorten >=2 ,scale = \thisplotscale, font=\footnotesize]
	
	\node[blue vertex] (x) {$x$} ;

	\path (x) ++ (4,0.7)   node [blue vertex]    (x1)   {$x_1$} 
	          ++ (4,0)   node [phantom vertex] (x2)   {$\ldots$}
	          ++ (0.5,0) node [phantom vertex] (xlm1) {$\ldots$} 
	          ++ (4,0)   node [blue vertex]    (xl)   {$x_{l-1}$}
	          ++ (4,-0.7)  node [blue vertex]    (xp)   {$x'$}; 

	\path (x) ++ (4,-1)  node [blue vertex]    (xlp)   {$x'_{l'-1}$} 
	          ++ (4,0)   node [phantom vertex] (xlm1p) {$\ldots$}
	          ++ (0.5,0) node [phantom vertex] (x2p)   {$\ldots$} 
	          ++ (4,0)   node [blue vertex]    (x1p)   {$x'_{1}$}; 

	\path (x)    edge [bend left, above] node {$A_X(x,x_1)$}       (x1);	
	\path (x1)   edge [bend left, above] node {$A_X(x_1,x_2)$}     (x2);	
	\path (xlm1) edge [bend left, above] node {$A_X(x_{l-2},x_{l-1})$} (xl);	
	\path (xl)   edge [bend left, above] node {$A_X(x_{l-1},x')$}      (xp);	

	\path (xp)    edge [bend left, below] node {$A_X(x',x'_1)$}       (x1p);
	\path (x1p)   edge [bend left, below] node {$A_X(x'_1,x'_2)$}     (x2p);
	\path (xlm1p) edge [bend left, below] node {$A_X(x'_{l'-2},x'_{l'-1})$} (xlp);
	\path (xlp)   edge [bend left, below] node {$A_X(x'_{l'-1},x)$}       (x);

\end{tikzpicture}
}

%% file: figures/reciprocal_nonreciprocal_dendrograms_2.tex
\def \thisplotscale {0.85}
\def \unit {\thisplotscale cm}

{\small
\begin{tikzpicture}[scale = \thisplotscale, x=1.0*\unit, y = 0.8*\unit]

    \path (6.5,-4) ++ (0, 0.0) node [blue vertex, scale=0.7] (1) {\large{$a$}}
                     +  (2, 0.0) node [blue vertex, scale=0.7] (2) {\large{$b$}}
                     +  (2,-5) node [blue vertex, scale=0.7] (3) {\large{$c$}}    
                     +  (0,-5) node [blue vertex, scale=0.7] (4) {\large{$d$}};

    \path (1) edge [bend left=20, above, -stealth, shorten >=2] node {$1$} (2);	
    \path (2) edge [bend left=20, right, -stealth, shorten >=2] node {$1$} (3);
    \path (3) edge [bend left=20, below, -stealth, shorten >=2] node {$1$} (4);    
    \path (4) edge [bend left=20, left,  -stealth, shorten >=2] node {$1$} (1);  	

    \path (2) edge [bend left=20, below, -stealth, shorten >=2] node {$3$} (1);	
    \path (3) edge [bend left=20, left,  -stealth, shorten >=2] node {$5$} (2);
    \path (4) edge [bend left=20, above, -stealth, shorten >=2] node {$2$} (3);   
    \path (1) edge [bend left=20, right, -stealth, shorten >=2] node {$5$} (4);

    \path [draw, -stealth] (-2,-9.5) 
           + (-0.5,   0) -- + (7,0.0) node [below, at end] {$\delta$};
    \path [draw, -stealth] (-2,-9.5)  
           + (   0,-0.5) -- + ( 0.0,7.5);
    \path [thin, draw=black!30]  (-2,-9.5) ++ (0,-0.2) 
                    ++ (1,0) node [below] {1} -- + (0,7.5)
                    ++ (1,0) node [below] {2} -- + (0,7.5)
                    ++ (1,0) node [below] {3} -- + (0,7.5)
                    ++ (1,0) 
                    ++ (1,0) node [below] {5} -- + (0,7.5)
                    ++ (1,0) node [below] {6} -- + (0,7.5);

    \path [draw, thick, draw=red] (-2,-9.5) ++ (0,5.5)
                      node [left] {$d$} -- +  (2, 0) 
           ++ (0,0.5) node [left] {$c$} -- ++ (2, 0) -- ++ (0,-0.5);
    \path [draw, thick, draw=blue](-2,-9.5) ++ (0,5.5)
           ++ (0,1.0) node [left] {$b$} -- +  (3, 0) 
           ++ (0,0.5) node [left] {$a$} -- ++ (3, 0) -- ++ (0,-0.5);
    \path [draw, thick, draw=black](-2,-9.5) ++ (0,5.5)
           ++ (0,0.25) ++ (2,0) -- +  (3, 0) 
           ++ (0,1.0)  ++ (1,0) -- ++ (2, 0) -- ++ (0,-1.0);
    \path [draw, thick, draw=black](-2,-9.5) ++ (0,5.5)
           ++ (0,0.75) ++ (5,0) -- +  (1.5, 0);
    \path (-2,-9.5) ++ (0,5.5) ++ (6.0,1.2) node [right] {$\ccalH^\R$};

    \path [draw, thick, draw=red] (-2,-9.5) ++ (0,3.0)
                      node [left] {$d$} -- +  (1, 0) 
           ++ (0,0.5) node [left] {$c$} -- ++ (1, 0) -- ++ (0,-0.5);
    \path [draw, thick, draw=blue](-2,-9.5) ++ (0,3.0)
           ++ (0,1.0) node [left] {$b$} -- +  (1, 0) 
           ++ (0,0.5) node [left] {$a$} -- ++ (1, 0) -- ++ (0,-0.5);
    \path [draw, thick, draw=black](-2,-9.5) ++ (0,3.0)
           ++ (0,1.0)  ++ (1,0) -- ++ (0,-0.5);
    \path [draw, thick, draw=black](-2,-9.5) ++ (0,3.0)
           ++ (0,0.75) ++ (1,0) -- +  (5.5, 0);
    \path (-2,-9.5) ++ (0,3.0) ++ (6.0,1.2) node [right] {$\ccalH^{\NR}$};

    \path [draw, thick, draw=red] (-2,-9.5) ++ (0,0.5)
                      node [left] {$d$} -- +  (1, 0) 
           ++ (0,0.5) node [left] {$c$} -- ++ (1, 0) -- ++ (0,-0.5);
    \path [draw, thick, draw=blue](-2,-9.5) ++ (0,0.5)
           ++ (0,1.0) node [left] {$b$} -- +  (1, 0) 
           ++ (0,0.5) node [left] {$a$} -- ++ (1, 0) -- ++ (0,-0.5);
    \path [draw, thick, draw=black](-2,-9.5) ++ (0,0.5)
           ++ (0,0.25) ++ (1,0) -- +  (4, 0) 
           ++ (0,1.0)  ++ (0,0) -- ++ (4, 0) -- ++ (0,-1.0);
    \path [draw, thick, draw=black](-2,-9.5) ++ (0,0.5)
           ++ (0,0.75) ++ (5,0) -- +  (1.5, 0);
    \path (-2,-9.5) ++ (0,0.5) ++ (6.0,1.2) node [right] {$\ccalH^{\R/\NR}$};
           
    \path [thick, draw=mygreen]  (-2,-9.5) ++ (0,-0.2) 
                    ++ (4,0) node [below] {$\beta=4$} -- + (0,7.5);

\end{tikzpicture}}

%% file: figures/inter_reciprocal_example.tex
\def \thisplotscale {1.2}
\def \unit {\thisplotscale cm}
\tikzstyle {blue vertex here} = [blue vertex, 
                                 minimum width = 0.4*\unit, 
                                 minimum height = 0.4*\unit, 
                                 anchor=center]
\tikzstyle {blue vertex here 2} = [blue vertex, 
                                 minimum width = 0.4*\unit, 
                                 minimum height = 0.4*\unit,
                                 font=\scriptsize, 
                                 anchor=center]
{\small \begin{tikzpicture}[thick, x = 1.2*\unit, y = 0.96*\unit]


	\path[draw, thin] (8.0,3.4) ++ ( -4, -0.5) node[blue vertex here] (1) {{$x$}}
	                            ++ ( 0.6, 0.5) node[blue vertex here 2] (2) {{$y_{01}$}}
	                            ++ (0.4,0)   node [phantom vertex] ()   {$\ldots$}
	                            ++ (0.18,0)   node [phantom vertex] ()   {$\ldots$}
	                            ++ (0.5,0) node[blue vertex here 2] (3) {{$y_{0k_0}$}}
	                            ++ (0.6, -0.5) node[blue vertex here] (4) {{$x_1$}}
	                            ++ (0.5, 0.4) node (p1) {}
	                            ++ (0, -0.4) node () {$\ldots$}
	                            ++ (0.18, 0) node () {$\ldots$}

	                            ++ (0, 0.4) node (p2) {}

	                            ++ ( 0.5, -0.4) node[blue vertex here] (5) {{$x_{r}$}}
	                            ++ ( 0.6, 0.5) node[blue vertex here 2] (9) {{$y_{r1}$}}
	                            ++ (0.4,0)   node [phantom vertex] ()   {$\ldots$}
	                            ++ (0.18,0)   node [phantom vertex] ()   {$\ldots$}
	                            ++ (0.5,0) node[blue vertex here 2] (10) {{$y_{rk_r}$}}
	                            ++ (0.6, -0.5) node[blue vertex here] (11) {{$x'$}}
                                     ++ (-0.6, -0.5) node[blue vertex here 2] (12) {{$y'_{r1}$}}
                                     ++ (-0.4,0)   node [phantom vertex] ()   {$\ldots$}
	                            ++ (-0.18,0)   node [phantom vertex] ()   {$\ldots$}
	                            ++ (-0.5, 0) node[blue vertex here 2] (13) {{$y'_{rk'_r}$}}
	                             ++ (-1.1, 0.1) node (p3) {}
	                             ++ (-0.22, 0) node (p4) {}


	                            ++ ( -1.1, -0.1) node[blue vertex here 2] (7) {{$y'_{01}$}}
	                             ++ (-0.4,0)   node [phantom vertex] ()   {$\ldots$}
	                            ++ (-0.18,0)   node [phantom vertex] ()   {$\ldots$} 
	                            ++ ( -0.48, 0) node[blue vertex here 2] (8) {{$y'_{0k'_0}$}};
    \path[thin, -stealth] (1) edge [bend left=10, above] node {} (2);		                            
    \path[thin, -stealth] (3) edge [bend left=10, below] node {} (4);
    \path[thin, -stealth] (4) edge [bend left=10, below] node {} (p1);	
    \path[thin, -stealth] (p2) edge [bend left=10, below] node {} (5);
    \path[thin, -stealth] (4) edge [bend left=10, above] node {} (7);	
    \path[thin, -stealth] (8) edge [bend left=10, right] node {} (1);	
    \path[thin, -stealth] (5) edge [bend left=10, right] node {} (9);	
    \path[thin, -stealth] (10) edge [bend left=10, right] node {} (11);
    \path[thin, -stealth] (5) edge [bend left=10, right] node {} (p3);	    
       \path[thin, -stealth] (p4) edge [bend left=10, right] node {} (4);   
       \path[thin, -stealth] (11) edge [bend left=10, right] node {} (12); 
       \path[thin, -stealth] (13) edge [bend left=10, right] node {} (5);  
\end{tikzpicture}}

%% file: figures/inter_reciprocal_example_2.tex
\def \thisplotscale {1.2}
\def \unit {\thisplotscale cm}
\tikzstyle {blue vertex here} = [blue vertex, 
                                 minimum width = 0.4*\unit, 
                                 minimum height = 0.4*\unit, 
                                 anchor=center]
{\small \begin{tikzpicture}[thick, x = 1.2*\unit, y = 0.96*\unit]
         
    \scriptsize

	\path[draw, thin] (8.0,3.4) ++ ( -6, -1) node[blue vertex here] (1) {{$x$}}
	                            ++ ( 1.0, 0.7) node[blue vertex here] (2) {{$x_1$}}
	                            ++ (1.5,0) node[blue vertex here] (3) {{$x_2$}}
	                            ++ ( 1, -.7) node[blue vertex here] (4) {{$x_3$}}
	                            ++ ( 1, .7) node[blue vertex here] (5) {{$x_4$}}
	                            ++ ( 1, -.7) node[blue vertex here] (6) {{$x'$}}
	                            ++ ( -1, -.7) node[blue vertex here] (7) {{$x_5$}}
	                            ++ ( -2.75, 0) node[blue vertex here] (8) {{$x_6$}};
    \path[thin, -stealth] (1) edge [bend left=15, above] node {$1$} (2);		                            
    \path[thin, -stealth] (2) edge [right, above] node {$1$} (3);
    \path[thin, -stealth] (3) edge [above] node {$1$} (5);	
    \path[thin, -stealth] (4) edge [bend left =10, above left, pos=0.4]  node {$2$} (5);	        
    \path[thin, -stealth] (5) edge [bend left=15, above] node {$1$} (6);		                            
    \path[thin, -stealth] (6) edge [bend left=15, below]  node {$1$} (7);	
    \path[thin, -stealth] (7) edge [below] node {$1$} (8);	
    \path[thin, -stealth] (8) edge [bend left = 15, below] node {$1$} (1);
    \path[thin, -stealth] (2) edge [bend left = 5, below, pos=0.1] node {$3$} (4);
    \path[thin, -stealth] (3) edge [bend left = 15, above right, pos=0.65] node {$2$} (4);
     \path[thin, -stealth] (7) edge [bend left = 15, below left, pos=0.7] node {$2$} (4);	       
     \path[thin, -stealth] (4) edge [bend left = 20, below, pos=0.3] node {$4$} (6);
     \path[thin, -stealth] (6) edge [bend left = 20, above, pos=0.3] node {$4$} (4); 
     \path[thin, -stealth] (1) edge [bend left = 10, below, pos=0.3] node {$4$} (4); 
     \path[thin, -stealth] (4) edge [bend left = 10, above, pos=0.3] node {$4$} (1); 
     \path[thin, -stealth] (4) edge [bend left = 10, below, pos=0.25] node {$2$} (8); 
\end{tikzpicture}}

%% file: figures/reciprocal_example_io_small.tex
\def \thisplotscale {1}
\def \unit {\thisplotscale cm}

{\footnotesize \begin{tikzpicture}[thick, x = 1*\unit, y = 1*\unit]

\node (dendrogram) at (0,0)
    {\centering\includegraphics [width=0.8\linewidth, height = 0.45\linewidth ]
    {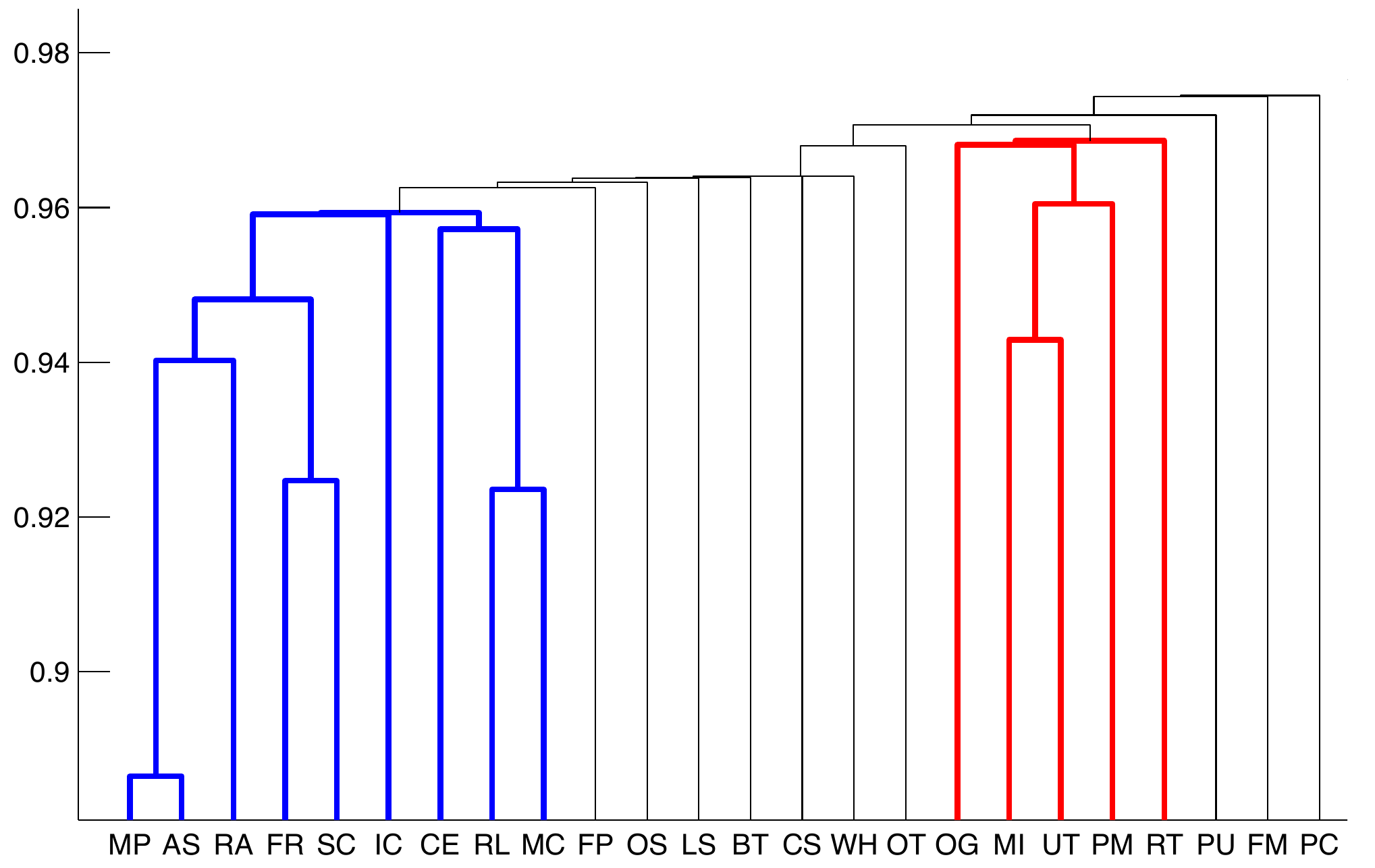}};      

\path (dendrogram) ++ (0, -3.5) node (cluster blue)
   {\centering\input{figures/reciprocal_io_network_1_small.tex}};

\path (dendrogram) ++ (0, -2.2) node {(a)};

\path (cluster blue)  ++(0,-1) node {$\delta^\R_1=0.959$};

\path (dendrogram) ++ (0,-5) node {(b)};

\end{tikzpicture}}

%% file: figures/reciprocal_io_network_1_small.tex
\def \thisplotscale {1}
\def \unit {\thisplotscale cm}
\tikzstyle {blue vertex here} = [blue vertex, 
                                 minimum width =  0.5*\unit, 
                                 minimum height = 0.5*\unit, 
                                 anchor=center]

{\scriptsize \begin{tikzpicture}[thick, x = 1*\unit, y = 1*\unit]
         
	\path[draw, thin]  
	      ( 0, 0) node[blue vertex here] (AS) {{AS}}
	      ( 0, 1) node[blue vertex here] (RA) {{RA}}
         ( 1, 0) node[blue vertex here] (MC) {{MC}}
	      (-1, 0) node[blue vertex here] (MP) {{MP}};

	\path[draw, thin] (MP) 
      ++(-1, 0) node[blue vertex here] (FR) {{FR}}
      ++(-1, 0) node[blue vertex here] (SC) {{SC}}
      ++(-1, 0) node[blue vertex here] (IC) {{IC}};

	\path[draw, thin] (MC) 
      + (0.707106, 0.707106) node[blue vertex here] (CE) {{CE}}
      + (0.707106,-0.707106) node[blue vertex here] (RL) {{RL}};

   \path[thin, stealth-stealth] (MP) edge (AS);	
   \path[thin, stealth-stealth] (MP) edge (FR);
   \path[thin, stealth-stealth] (AS) edge (RA);
   \path[thin, stealth-stealth] (AS) edge (MC);
   \path[thin, stealth-stealth] (SC) edge (FR);
   \path[thin, stealth-stealth] (SC) edge (IC);
   \path[thin, stealth-stealth] (CE) edge (MC);
   \path[thin, stealth-stealth] (MC) edge (RL);	                            
   
\end{tikzpicture}}

%% file: figures/nonreciprocal_example_io_small.tex
\def \thisplotscale {1}
\def \unit {\thisplotscale cm}

{\footnotesize \begin{tikzpicture}[thick, x = 1*\unit, y = 1*\unit]

\node (dendrogram) at (0,0)
    {\centering\includegraphics [width=0.8\linewidth, height = 0.4\linewidth]
    {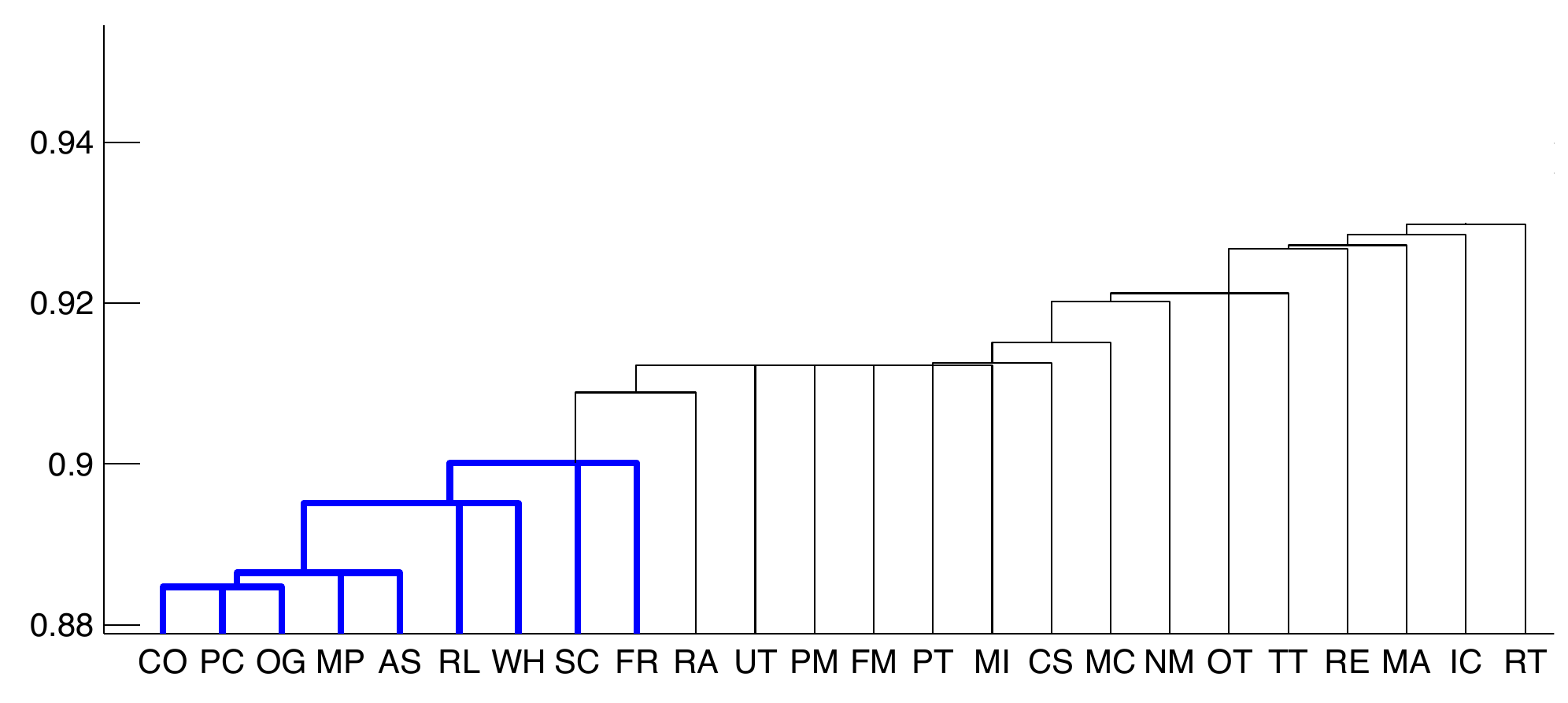}};      

\path (dendrogram) ++ (0,-3.8) node (cluster 1)
   {\centering\input{figures/nonreciprocal_io_network_4_small.tex}};      

\path (dendrogram) ++ (0, -1.9) node {(a)};

\path (cluster 1)  ++(-2.5,-0.75) node {$\delta^{\NR}_1=0.900$};

\path (dendrogram) ++ (0,-5.5) node {(b)};

\end{tikzpicture}}

%% file: figures/nonreciprocal_io_network_4_small.tex
\def \thisplotscale {0.95}
\def \unit {\thisplotscale cm}
\tikzstyle {blue vertex here} = [blue vertex, thin,
                                 minimum width = 0.5*\unit, 
                                 minimum height = 0.5*\unit, 
                                 anchor=center]
{\scriptsize \begin{tikzpicture}[thick, x = 1.4*\unit, y = 1.2*\unit]

	\node[blue vertex here] (PC) {{PC}};
	
	\path (PC)
      + ( 0,     -1)      node [blue vertex here] (CO) {{CO}}
      + (-0.8661,-0.5) node [blue vertex here] (OG) {{OG}}
      + ( 1,      0)      node [blue vertex here] (AS) {{AS}}
      + ( 1,     -1)      node [blue vertex here] (MP) {{MP}}
      + ( 1,-1.8) node [blue vertex here] (WH) {{WH}}
      + ( 0,-1.8) node [blue vertex here] (RL) {{RL}}
      + ( 2,-0.5) node [blue vertex here] (SC) {{SC}}
      + ( 2,-1.5) node [blue vertex here] (FR) {{FR}};

  \path[thin, -stealth] (PC) edge  node {} (CO);	
  \path[thin, -stealth] (CO) edge [bend left] node {} (OG);
  \path[very thick, -stealth] (OG) edge [bend left] node {} (PC);
  \path[very thick, -stealth] (AS) edge [bend left] node {} (MP);
  \path[thin, -stealth] (MP) edge [bend left] node {} (AS);
  \path[very thick, -stealth] (PC) edge [bend left] node {} (AS);
  \path[thin, -stealth] (MP) edge [bend left] node {} (CO);
  \path[thin, -stealth] (MP) edge [bend left] node {} (RL);
  \path[thin, -stealth] (MP) edge [bend left=20] node {} (WH);
  \path[thin, -stealth] (WH) edge [bend left=20] node {} (RL);
  \path[very thick, -stealth] (RL) edge [bend left] node {} (OG);
    \path[very thick, -stealth] (MP) edge [bend left] node {} (SC);
      \path[thin, -stealth] (MP) edge [bend left] node {} (FR);
        \path[very thick, -stealth] (SC) edge [bend left] node {} (FR);
          \path[very thick, -stealth] (FR) edge [bend left=40] node {} (RL);

\end{tikzpicture}}

%% file: figures/semi_reciprocal_example_io_small.tex
\def \thisplotscale {1}
\def \unit {\thisplotscale cm}

{\footnotesize \begin{tikzpicture}[thick, x = 1*\unit, y = 1*\unit]

\node (dendrogram) at (0,0)
    {\centering\includegraphics [width=0.8\linewidth, height = 0.45\linewidth ]
        {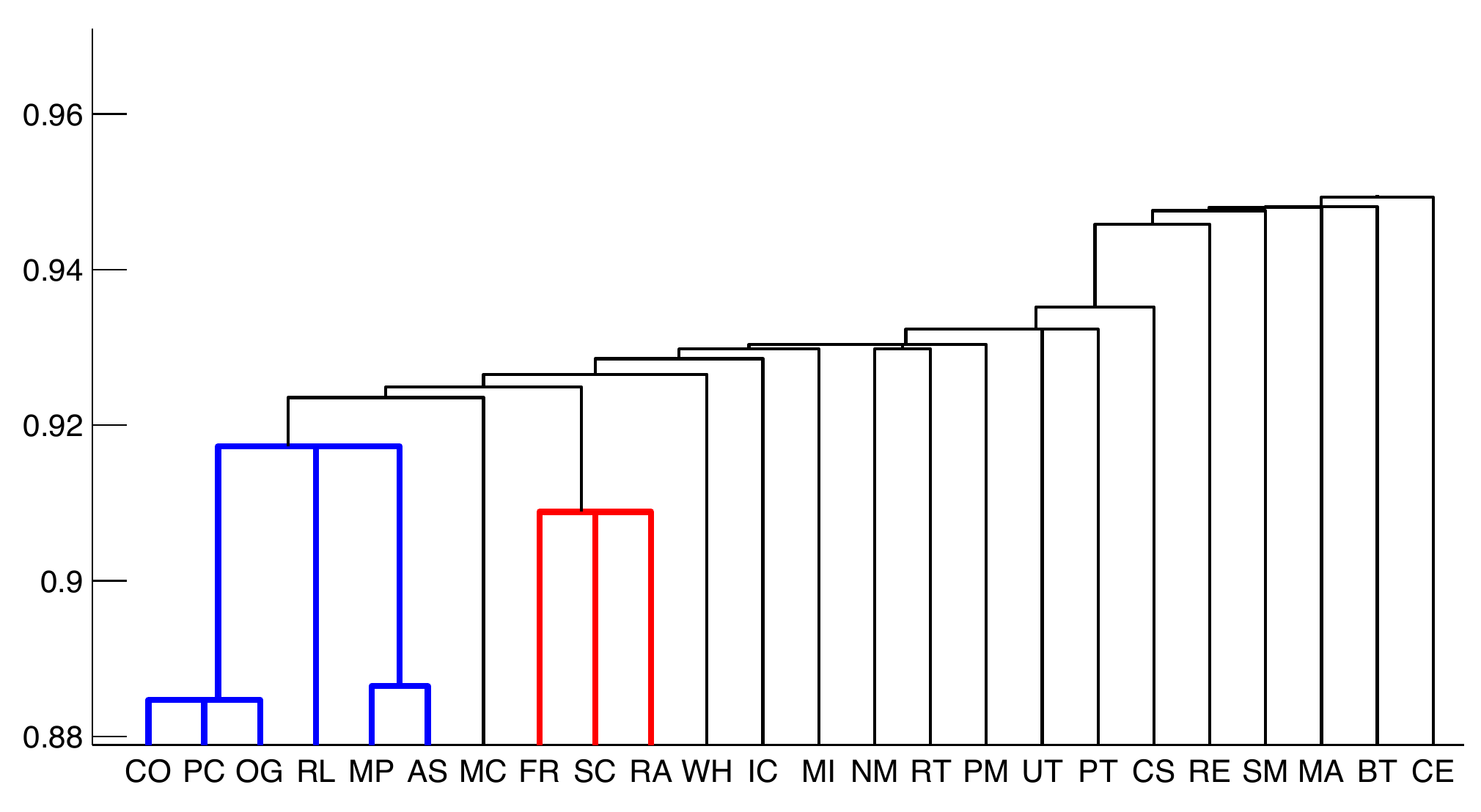}};      

\path (dendrogram) ++ (-2.4,-3.7) node (cluster 1)
   {\centering\input{figures/semi_reciprocal_io_network_2_small.tex}};      

\path (dendrogram) ++ (1.4,-3.7) node (cluster 2)
   {\centering\input{figures/semi_reciprocal_io_network_1_small.tex}};      

\path (dendrogram) ++ (0, -2.2) node {(a)};

\path (cluster 1)  ++(0,-1.3) node {$\delta^\SR_1=0.909$};
\path (cluster 2)  ++(0.2,-1.3) node {$\delta^\SR_2=0.917$};
\path (dendrogram) ++ (0,-5.5) node {(b)};

\end{tikzpicture}}

%% file: figures/semi_reciprocal_io_network_2_small.tex
\def \thisplotscale {1}
\def \unit {\thisplotscale cm}
\tikzstyle {red vertex here} = [red vertex, thin,
                                 minimum width = 0.5*\unit, 
                                 minimum height = 0.5*\unit, 
                                 anchor=center]
{\scriptsize \begin{tikzpicture}[thick, x = 1.4*\unit, y = 1.2*\unit]

	\node[red vertex here] (SC) {{SC}};
	
	\path (SC)
      + ( 0,     -1)      node [red vertex here] (RA) {{RA}}
      + (-0.8661,-0.5) node [red vertex here] (FR) {{FR}};
	                           
  \path[very thick, -stealth] (RA) edge  node {} (SC);	
  \path[very thick, -stealth] (SC) edge [bend right] node {} (FR);
  \path[very thick, -stealth] (FR) edge [bend right] node {} (RA);
        
\end{tikzpicture}}

%% file: figures/semi_reciprocal_io_network_1_small.tex
\def \thisplotscale {1}
\def \unit {\thisplotscale cm}
\tikzstyle {blue vertex here} = [blue vertex, thin,
                                 minimum width = 0.5*\unit, 
                                 minimum height = 0.5*\unit, 
                                 anchor=center]
{\scriptsize \begin{tikzpicture}[thick, x = 1.4*\unit, y = 1.2*\unit]

	\node[blue vertex here] (PC) {{PC}};
	
	\path (PC)
      + ( 0,     -1)      node [blue vertex here] (OG) {{OG}}
      + (-0.8661,-0.5) node [blue vertex here] (CO) {{CO}}
      + (0.8661, -0.5) node [blue vertex here] (RL) {{RL}}
      + (2*0.8661, 0) node [blue vertex here] (AS) {{AS}}
      + (2*0.8661, -1) node [blue vertex here] (MP) {{MP}};
	                           
  \path[very thick, -stealth] (PC) edge [bend right] node {} (CO);	
  \path[very thick, -stealth] (CO) edge [bend right] node {} (OG);
  \path[very thick, -stealth] (OG) edge node {} (PC);
  \path[very thick, -stealth] (RL) edge [bend left] node {} (OG);
  \path[very thick, -stealth] (AS) edge [bend right] node {} (RL);
  \path[thin, -stealth] (MP) edge [bend left] node {} (RL);
  \path[very thick, -stealth] (AS) edge [bend right] node {} (MP);
  \path[very thick, -stealth] (MP) edge [bend right] node {} (AS);
  \path[thin, -stealth] (MP) edge [bend left=55] node {} (CO);
  \path[very thick, -stealth] (PC) edge [bend left] node {} (AS);
        
\end{tikzpicture}}